\documentclass{article}

\usepackage{amsmath,amsfonts,bm}

\def\eqref#1{equation~\ref{#1}}

\def\1{\bm{1}}

\def\va{{\bm{a}}}
\def\vb{{\bm{b}}}

\def\vh{{\bm{h}}}

\def\vv{{\bm{v}}}
\def\vw{{\bm{w}}}
\def\vx{{\bm{x}}}

\def\mA{{\bm{A}}}

\def\mU{{\bm{U}}}

\DeclareMathAlphabet{\mathsfit}{\encodingdefault}{\sfdefault}{m}{sl}
\SetMathAlphabet{\mathsfit}{bold}{\encodingdefault}{\sfdefault}{bx}{n}

\usepackage{ulem}

\usepackage{microtype}
\usepackage{graphicx}
\usepackage{subfigure}

\PassOptionsToPackage{numbers, compress}{natbib}
\usepackage[final]{neurips_2021}

\usepackage[utf8]{inputenc} 
\usepackage[T1]{fontenc}    
\usepackage{hyperref}       
\usepackage{url}            
\usepackage{booktabs}       
\usepackage{amsfonts}       
\usepackage{nicefrac}       
\usepackage{microtype}      
\usepackage{xcolor}         

\newcommand{\emdash}[1]{\,---\,#1\,---\,}

\title{Reverse engineering learned optimizers\\reveals known and novel mechanisms}

\author{%
  Niru Maheswaranathan\thanks{Work conducted while at Google Research. Currently at Meta Reality Labs.} \\
  Google Research, Brain Team\\
  \texttt{niru@hey.com} \\
  \And
  David Sussillo$^*$ \\
  Google Research, Brain Team \\
   \AND
  Luke Metz \\
  Google Research, Brain Team \\
     \texttt{lmetz@google.com} \\
   \And
   Ruoxi Sun \\
   Google Research, Brain Team \\
    \texttt{ruoxis@google.com} \\
   \And
   Jascha Sohl-Dickstein \\
   Google Research, Brain Team \\
   \texttt{jaschasd@google.com} \\
   \setcounter{footnote}{0}
}

\begin{document}

\maketitle

\begin{abstract}
Learned optimizers are parametric algorithms that can themselves be trained to solve optimization problems. In contrast to baseline optimizers (such as momentum or Adam) that use simple update rules derived from theoretical principles, learned optimizers use flexible, high-dimensional, nonlinear parameterizations. Although this can lead to better performance, their inner workings remain a mystery. How is a given learned optimizer able to outperform a well tuned baseline? Has it learned a sophisticated combination of existing optimization techniques, or is it implementing completely new behavior? In this work, we address these questions by careful analysis and visualization of learned optimizers. We study learned optimizers trained from scratch on four disparate tasks, and discover that they have learned interpretable behavior, including: momentum, gradient clipping, learning rate schedules, and learning rate adaptation. Moreover, we show how dynamics and mechanisms inside of learned optimizers orchestrate these computations. Our results help elucidate the previously murky understanding of how learned optimizers work, and establish tools for interpreting future learned optimizers.
\end{abstract}

\section{Introduction}
\label{sec:introduction}
Optimization algorithms underlie nearly all of modern machine learning; thus advances in optimization have broad impact.
Recent research uses meta-learning to learn new optimization algorithms, by directly parameterizing and training an optimizer on a distribution of tasks.
These so-called \textit{learned optimizers} have been shown to outperform baseline optimizers in restricted settings \citep{andrychowicz2016learning,wichrowska2017learned,lv2017learning,bello2017neural,li2016learning,metz2019understanding,metz2020tasks}.

Despite improvements in the design, training, and performance of learned optimizers, fundamental questions remain about their behavior.
We understand remarkably little about \textit{how} these optimizers work.
Are learned optimizers simply learning a clever combination of known techniques?
Or do they learn fundamentally new behaviors that have not yet been proposed in the optimization literature?
If they did learn a new optimization technique, how would we know?

Contrast this with existing ``hand-designed'' optimizers such as momentum \citep{polyak1964some}, AdaGrad \citep{duchi2011adaptive}, RMSProp \citep{tieleman2012lecture}, or Adam \citep{kingma2014adam}.
These algorithms are motivated and analyzed using intuitive mechanisms and theoretical principles (such as accumulating update velocity in momentum, or rescaling updates based on gradient magnitudes in RMSProp or Adam).
This understanding of underlying mechanisms allows future studies to build on these techniques by highlighting flaws in their operation \citep{loshchilov2018fixing}, studying convergence \citep{reddi2019convergence}, and developing deeper knowledge about why key mechanisms work \citep{Zhang2020Why}. Without analogous 
understanding of the inner workings of a learned optimizers, it is incredibly difficult to analyze or synthesize their behavior.

In this work, we develop tools for isolating and elucidating mechanisms in nonlinear, high-dimensional learned optimization algorithms (\S\ref{sec:tools}).
Using these methods we show how learned optimizers utilize both known and novel techniques, across four disparate tasks.
In particular, we demonstrate that learned optimizers learn momentum (\S\ref{sec:momentum}), gradient clipping (\S\ref{sec:clipping}), learning rate schedules (\S\ref{sec:schedules}), and methods for learning rate adaptation (\S\ref{sec:adaptation}, \S\ref{sec:systemid}).
Taken together, our work can be seen as part of a new approach to scientifically interpret and understand learned algorithms.

We provide code for training and analyzing learned optimizers, as well as the trained weights for the learned optimizers studied here, at \texttt{\url{https://bit.ly/3eqgNrH}}.

\section{Related Work}

Our work is heavily inspired by recent work using neural networks to parameterize optimizers. \citet{andrychowicz2016learning} originally showed promising results on this front, with additional studies improving robustness~\citep{wichrowska2017learned,lv2017learning}, meta-training~\citep{metz2019understanding}, and generalization~\citep{metz2020tasks} of learned optimizers.

Here, we study the behavior of optimizers by treating them as dynamical systems. This perspective has yielded a number of intuitive and theoretical insights~\citep{su2014differential,wilson2016lyapunov,goh2017momentum,shi2019acceleration}. We also build on recent work on reverse engineering recurrent neural networks (RNNs).
\citet{sussillo2013opening} showed how linear approximations of nonlinear RNNs can reveal the algorithms used by trained networks to solve simple tasks. These techniques have been applied to understand trained RNNs in a variety of domains, from natural langauge processing~\citep{maheswaranathan2019reverse,maheswaranathan2020recurrent} to neuroscience~\citep{schaeffer2020reverse}. Additional work on treating RNNs as dynamical systems has led to insights into their computational capabilities~\citep{jordan2019gated,krishnamurthy2020theory,can2020gating}.

\section{Methods}
\label{sec:methods}

\subsection{Preliminaries}
We are interested in optimization problems that minimize a loss function ($f$) over parameters ($\vx$). We focus on first-order optimizers, which at iteration $k$ have access to the gradient $g_i^k \equiv \nabla f(x_i^k)$ and produce an update $\Delta x_i^k$. These are \textit{component-wise} optimizers that are applied to each parameter ($x_i$) of the problem in parallel. Standard optimizers used in machine learning (e.g. momentum, Adam) are in this category\footnote{Notable exceptions include quasi-Newton methods such as L-BFGS~\citep{nocedal2006numerical} or K-FAC~\citep{martens2015optimizing}.}.
Going forward, we use $x$ for the parameter to optimize, $g$ for its gradient, $k$ for the current iteration, and drop the parameter index ($i$) to reduce excess notation.

One can think of an optimizer as being comprised of two parts: the optimizer state ($\vh$) that stores information about the current problem, and readout weights ($\vw$) that are used to update parameters given the current state. An optimization algorithm is specified by an initial state, state transition dynamics, and readout, defined as follows:
\begin{eqnarray}
    \vh^{k+1} &=& F(\vh^k, g^k) \label{eq:dynamics} \\
    x^{k+1} &=& x^{k} + \vw^T\vh^{k+1}, \label{eqn:readout}
\end{eqnarray}
where $\vh$ is the optimizer state, $F$ governs the optimizer dynamics, and $\vw$ are the readout weights.

\textit{Learned optimizers} are constructed by parameterizing the function $F$, and then learning those parameters along with the readout weights through meta-optimization (detailed in App.~\ref{appendix:metatraining}).
\textit{Hand-designed} optimization algorithms, by distinction, specify these functions at the outset.

For example, in momentum, the state variable is a single number (known as the velocity), and is updated as a linear combination of the previous state and the gradient (e.g.~$h^{k+1} = \beta h^k + g^k$, where $\beta$ is the momentum hyperparameter).
For momentum and other hand-designed optimizers, the state variables are low-dimensional, and their dynamics are (largely) straightforward. In contrast, learned optimizers have high-dimensional state variables, and the potential for rich, nonlinear dynamics. As these systems learn complex behaviors, it has historically been difficult to extract simple, intuitive descriptions of the behavior of a learned optimizer.

\begin{figure*}
\centering
\includegraphics[width=1.0\textwidth]{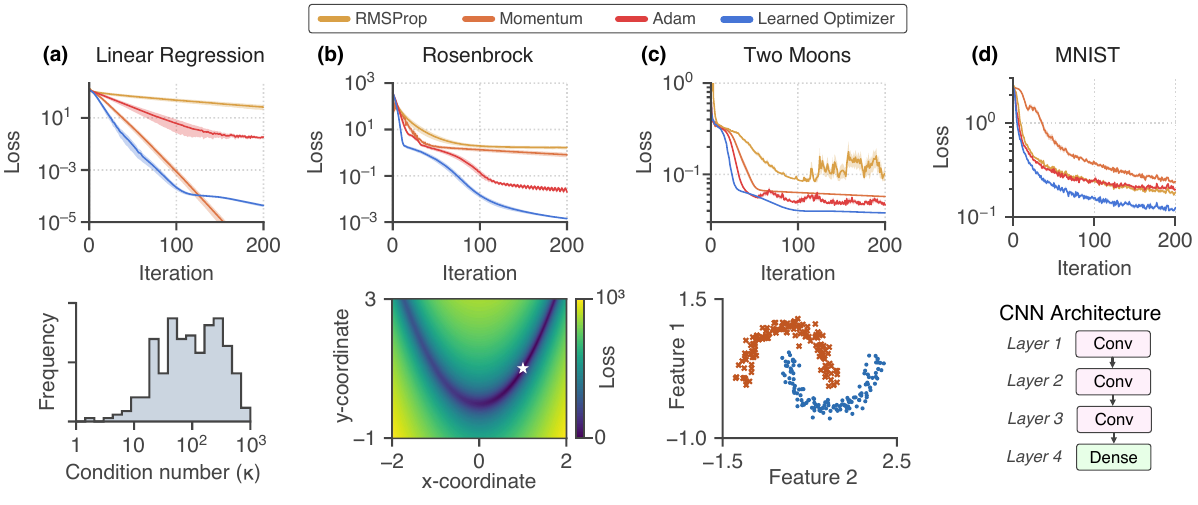}
\vspace*{-5mm}
\caption{Learned optimizers outperform well tuned baselines on four tasks: \textbf{(a)} linear regression, \textbf{(b)} the Rosenbrock function, \textbf{(c)} training a fully connected neural network on the two moons dataset, and \textbf{(d)} training a convolutional neural network on the MNIST dataset. \textit{Top row}: Optimizer performance, shown as loss curves (mean $\pm$ std. error over 64 random seeds) for momentum (orange), RMSProp (yellow), Adam (red) and a learned optimizer (blue). \textit{Bottom row}: Additional information pertaining to each task (described in \S\ref{sec:training}).}
\label{fig:perf}
\end{figure*}

\subsection{Training learned optimizers}
\label{sec:training}

We parametrize the learned optimizer with a recurrent neural network (RNN), similar to \citet{andrychowicz2016learning}.
The only input to the optimizer is the gradient. The RNN is trained by minimizing a meta-objective, which we define as the average training loss when optimizing a target problem. See App.~\ref{appendix:metatraining} for details about the optimizer architecture and meta-training procedures. Below, we only analyze the final (best) trained optimizer, however we do analyze aspects of the meta-training dynamics in App~\ref{appendix:metatrainingdynamics}.

We trained learned optimizers on each of four tasks. These tasks were selected because they converge in a relatively small number of iterations (particularly important for meta-optimization) and cover a range of loss surfaces (convex and non-convex, low- and high-dimensional, deterministic and stochastic):

\textbf{Linear regression:} The first task consists of random linear regression problems $f(\vx) = \frac{1}{2}\|\mA \vx - \vb\|_2^2$, where $\mA$ and $\vb$ are randomly sampled. Much of our theoretical understanding of the behavior of optimization algorithms is derived using quadratic loss surfaces such as these, in part because they have a constant Hessian ($\mA^T\mA$) over the entire parameter space. The choice of how to sample the problem data $\mA$ and $\vb$ will generate a particular distribution of Hessians and condition numbers. A histogram of condition numbers for our task distribution is shown in Figure~\ref{fig:perf}a.

\textbf{Rosenbrock:} The second task is minimizing the Rosenbrock function \citep{rosenbrock1960automatic}, a commonly used test function for optimization. It is a non-convex function with a curved valley and a single global minimum. The function is defined over two parameters as $f(x,y) = (1 - x)^2 + 100(y-x^2)^2$ (Figure~\ref{fig:perf}b). The distribution of problems for this task consists of different initializations sampled uniformly over a grid. The grid used to sample initializations is the same as the grid shown in the figure; the x- and y- coordinates are sampled from the ranges (-2, 2) and (-1, 3), respectively.

\textbf{Two moons:} The third task involves training a fully connected neural network to classify a toy dataset, the two moons dataset (Figure~\ref{fig:perf}c).
As the data are not linearly separable, a nonlinear classifier is required.
The optimization problem is to train the weights of a three hidden layer fully connected neural network, with 64 units per layer and tanh nonlinearities (for a total of 8,577 parameters).
The distribution of problems involves sampling the initial weights of the fully connected network.

\textbf{MNIST:} The fourth task is to train a four layer convolutional network to classify digits from the MNIST dataset. We use a minibatch size of 100 examples; thus the gradients fed to the optimizer are stochastic, unlike the previous three problems. The network consists of three convolutional layers each with 16 channels with a 3$\times$3 kernel size and ReLU activations, followed by a final (fully connected) dense layer, for a total of 82,250 parameters (Figure~\ref{fig:perf}d).

We additionally tuned three baseline optimizers (momentum, RMSProp, and Adam) individually for each task. We selected the hyperparameters for each problem out of 2500 samples randomly drawn from a grid. Details about the exact grid ranges used for each task are in App.~\ref{appendix:baselines}.

Figure \ref{fig:perf} (top row) compares the performance of the learned optimizer (blue) to baseline optimizers (red, yellow, and orange), on each of the four tasks described above. Across all tasks, the learned optimizer outperforms the baseline optimizers on the meta-objective\footnote{As the meta-objective is the average training loss during an optimization run, it naturally penalizes the training curve earlier in training (when loss values are large). This explains the discrepancy in the curves for linear regression (Fig.~\ref{fig:perf}a, top) where momentum continues to decrease the loss late in training. Despite this, the learned optimizer has an overall smaller meta-objective due to having lower loss at earlier iterations.}  (App.~Fig.~\ref{fig:mobj}).
\section{Tools for understanding optimizers}
\label{sec:tools}
In order to analyze learned optimizers, we make extensive use of two methods. The first is a way to visualize what an optimizer is doing at a particular optimizer state. The second is a way of making sense of how the optimizer state changes, that is, understanding the optimizer state dynamics. We describe these below, and then use them to analyze optimizer behavior and mechanisms in \S\ref{sec:results}.

\subsection{Update functions}
\label{sec:updatefunction}

\begin{figure*}[h!]
\centering
\includegraphics[width=1.0\textwidth]{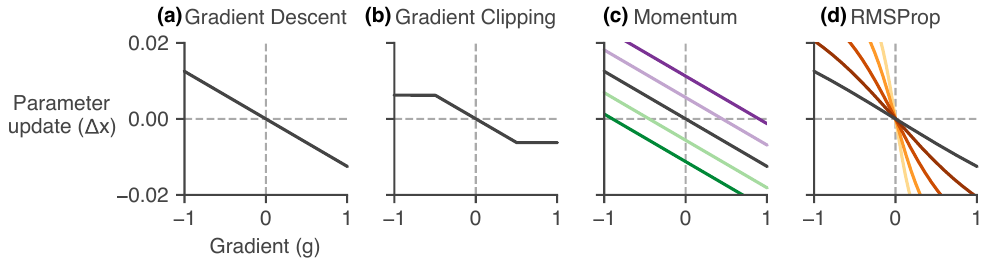}
\vspace{-2mm}
\caption{Visualizing optimizer behavior with update functions (see \S\ref{sec:updatefunction} for details) for different commonly used optimization techniques. 
\textbf{(a)} Gradient descent is a (stateless) linear function, whose slope is the learning rate.
\textbf{(b)} Gradient clipping saturates the update, beyond a threshold.
\textbf{(c)} Momentum introduces a vertical offset depending on the accumulated velocity (colors indicate different values of the accumulated momentum).
\textbf{(d)} RMSProp changes the slope (effective learning rate) of the update (colors denote changes in the state variable, the accumulated squared gradient).}
\label{fig:transfer}
\end{figure*}

First, we introduce a visualization tool to get a handle on what an optimizer is doing.
Any optimizer, at a particular state, can be viewed as a scalar function that takes in a gradient ($g$) and returns a change in the parameter ($\Delta x$). We refer to this as the optimizer \textit{update function}.

Mathematically, the update function is computed as the state update projected onto the readout, $\Delta x = \vw^TF(\vh, g)$, following equations (\ref{eq:dynamics}) and (\ref{eqn:readout}). In addition, the slope of this function with respect to the input gradient $\left(\frac{\partial \Delta x}{\partial g}\right)$ can be thought of as the \textit{effective learning rate} at a particular optimizer state\footnote{We compute this slope at $g=0$, in the middle of the update function. We find that the update function is always affine in the middle with saturation at the extremes, thus the slope at $g=0$ is a natural way to summarize the effective learning rate.}.
A steeper slope means that the parameter update is more sensitive to the gradient (as expected for a higher learning rate), and a shallower slope means that parameter updates are smaller for the same gradient magnitude (i.e.~a lower learning rate).

As the optimizer state varies, the update function and effective learning rate can change. This provides a mechanism for learned optimizers to implement different types of behavior: through optimizer state dynamics that induce particular types of changes in the update function.

It is instructive to first visualize update functions for commonly used optimizers (Figure~\ref{fig:transfer}). For gradient descent, the update ($\Delta x = -\alpha g$) is stateless and is always a fixed linear function whose slope is the learning rate, $\alpha$ (Fig.~\ref{fig:transfer}a).
Gradient clipping is also stateless, but is a saturating function of the gradient (Fig.~\ref{fig:transfer}b).
For momentum, the update is $\Delta x = \beta v - \alpha g$, where $v$ denotes the momentum state (velocity) and $\beta$ is the momentum timescale. The velocity adds an offset to the update function (Fig.~\ref{fig:transfer}c).
As the optimizer picks up positive (or negative) momentum, the curve shifts downward (or upward), thus incorporating a bias to reduce (or increase) the parameter.
For adaptive optimizers such as RMSProp, the state variable changes the slope, or effective learning rate, within the linear region of the update function (Fig.~\ref{fig:transfer}d).

Now, what about learned optimizers, or optimizers with much more complicated or high-dimensional state variables?
One advantage of update functions is that, as scalar functions, they can be easily visualized and compared to the known methods in Figure~\ref{fig:transfer}.
Whether or not the underlying hidden states are interpretable, for a given learned optimizer, remains to be seen.
 
\subsection{A dynamical systems perspective}
\label{sec:dynamicalsystems}

In order to understand the state dynamics of a learned optimizer, we approximate the nonlinear dynamical system (eq.~(\ref{eq:dynamics})) via linearized approximations~\citep{strogatz2018nonlinear}. These linear approximations hold near \textit{fixed points} of the dynamics.
Fixed points are points in the state space of the optimizer, where\emdash{as long as input gradients do not perturb it}the system does not move. That is, an approximate fixed point $\vh^*$ satisfies the following: $ \vh^* \approx F(\vh^*, g^*) $, for a particular input $g^*$.

We numerically find approximate fixed points \citep{sussillo2013opening,maheswaranathan2019universality} by solving an optimization problem where we find points ($\vh$) that minimize the following loss: $\frac{1}{2} \| F(\vh, g^*) - \vh \|_2^2$. The solutions to this problem (there may be many) are approximate fixed points of the system $F$ for a given input, $g^*$. In general, there may be different fixed points for different values of the input ($g$). First we will analyze fixed points when $g^*=0$ (\S\ref{sec:momentum}), and then later discuss additional behavior that occurs as $g^*$ varies (\S\ref{sec:adaptation}).

One can think of the structure of fixed points as shaping a dynamical skeleton that governs the optimizer behavior.
As we will see, for a well trained optimizer, the dynamics around fixed points enable interesting and useful computations.


\section{Mechanisms of learned optimizers}
\label{sec:results}
We selected and analyzed the best learned optimizer on each of the four tasks in \S\ref{sec:training}.
Across these, we discovered a number of mechanisms responsible for their superior performance: momentum, gradient clipping, learning rate schedules, and new types of learning rate adaptation
In the following sections, we go through each mechanism in detail, showing how it is implemented.

In general, we found similar mechanisms across learned optimizers trained on all four tasks. Thus, for brevity, we only show the results in the main text from one optimizer for each mechanism. For any mechanisms that were found to be task dependent, we point this out in the relevant section. Results for all tasks are presented in App.~\ref{appendix:results}.

\begin{figure*}[h]
\centering
\includegraphics[width=1.0\textwidth]{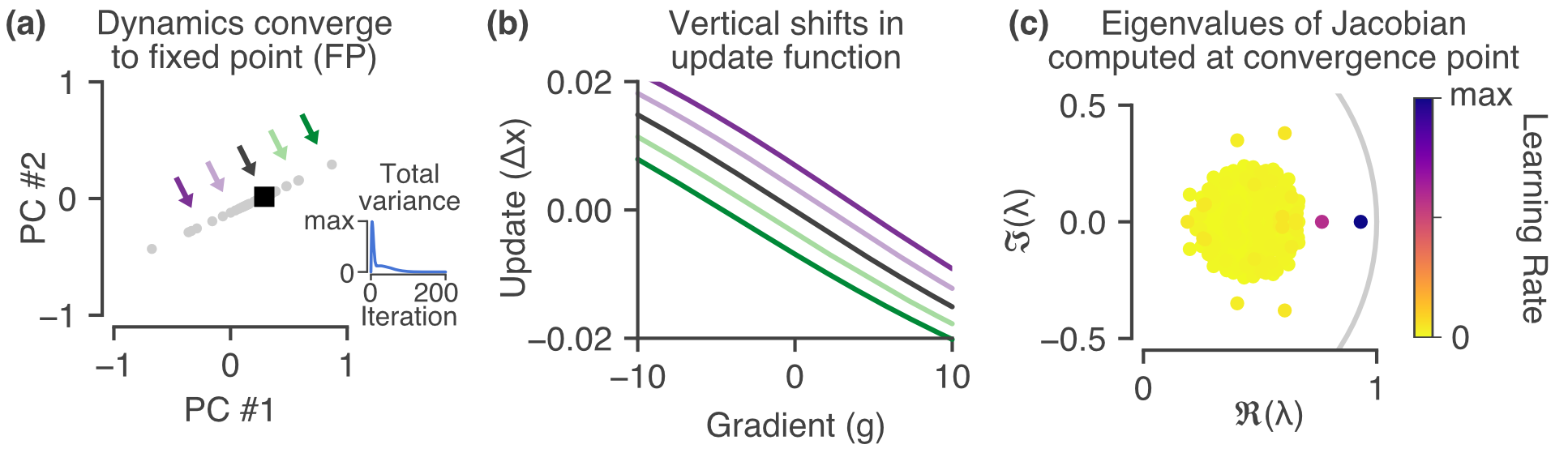}
\vspace{-2mm}
\caption{Momentum in learned optimizers. Plots are for the optimizer trained on the Rosenbrock task (we see similar behavior for optimizers trained on the other tasks, see App.~\ref{appendix:momentum}). \textbf{(a)}: Projection of the optimizer state near a convergence point (black square). {\it Inset:} the total variance of the optimizer states over test problems goes to zero as the trajectories converge. \textbf{(b)}: visualization of the update functions (\S\ref{sec:updatefunction}) along the slow mode of the dynamics (colored lines correspond to arrows in (a)). Along this dimension, the effect on the system is to induce an offset in the update, just as in classical momentum (cf. Fig.~\ref{fig:transfer}c). \textbf{(c)}: Eigenvalues of the linearized optimizer dynamics at the fixed point (black square in (a)) plotted in the complex plane. The eigenvalue magnitudes are momentum timescales, and the color indicates the corresponding learning rate. See \S\ref{sec:momentum} for details.}
\label{fig:momentum}
\end{figure*}

\subsection{Momentum}
\label{sec:momentum}

We discovered that learned optimizers implement classical momentum, and do so using dynamics that are well described by a linear approximation. 

To see how, first consider the dynamics of a learned optimizer near a fixed point. Here, we can linearly approximate the state dynamics (eq.~\ref{eq:dynamics}) using the Jacobian of the optimizer update~\citep{strogatz2018nonlinear,sussillo2013opening}. The \textit{linearized} state update is given by
\begin{eqnarray}
   F(\vh^k, g^k) \approx \vh^* + \frac{\partial F}{\partial \vh} \left(\vh^k - \vh^*\right) + \frac{\partial F}{\partial g} g^k,
\end{eqnarray}
where $\vh^*$ is a fixed point of the dynamics, $\frac{\partial F}{\partial \vh}$ is the Jacobian matrix, and $\frac{\partial F}{\partial g}$ is a vector that controls how the scalar gradient enters the system. Both of these latter two quantities are evaluated at the fixed point, $\vh^*$, and $g^*=0$.

This Jacobian is a matrix with $N$ eigenvalues and eigenvectors, where $N$ is the dimensionality of the optimizer state (for the RNN architectures that we use, $N$ is the number of RNN units). For a linear dynamical system, as we have now, the dynamics decouple along the $N$ eigenmodes of the system. 

Writing the update along these coordinates (let's denote them as $v_j$, for $j = 1, \ldots, N$) allows us to rewrite the learned optimizer as a momentum algorithm (see App.~\ref{appendix:aggmo} for details) with $N$ timescales:
$$v_j^{k+1} = \beta_j v_j^{k} + \alpha_j g + \text{const.},$$
where the magnitude of the eigenvalues of the Jacobian are exactly momentum timescales ($\beta_j$), each with a corresponding learning rate ($\alpha_j$). Incidentally, momentum with multiple timescales has been previously studied and called aggregated momentum by \citet{lucas2018aggregated}. 

To summarize, near a fixed point, a (nonlinear) learned optimizer is approximately linear, and the eigenvalues of the its Jacobian can be interpreted as momentum timescales.

We then looked to see if (and when) learned optimizers operated near fixed points. We found that across all tasks, optimizers converged to fixed points; often to a single fixed point. Figure~\ref{fig:momentum} shows this for a learned optimizer trained on the Rosenbrock task. Fig.~\ref{fig:momentum}a is a 2D projection of the hidden state (using principal components analysis\footnote{We run PCA across a large set of optimizer states visited during test examples to visualize optimizer state trajectories.}), and shows the single fixed point for this optimizer (black square). All optimizer state trajectories converge to this fixed point, this can be seen as the total variance of optimizer states across test examples goes to zero (Fig.~\ref{fig:momentum}a, inset).

Around this fixed point, the dynamics are organized along a line (gray circles). Shifting the hidden state along this line (indicated by colored arrows) induces a corresponding shift in the update function (Fig.~\ref{fig:momentum}b), similar to what is observed in classical momentum (cf.~Fig.~\ref{fig:transfer}c).

This learned optimizer uses a single eigenmode to implement momentum. Fig.~\ref{fig:momentum}c shows the eigenvalues of the Jacobian (computed at the convergence fixed point) in the complex plane, colored by that mode's learning rate (see App.~\ref{appendix:aggmo} for how these quantities are computed). This reveals a single dominant eigenmode (colored in purple), whose eigenvector corresponds to the momentum direction (gray points in Fig.~\ref{fig:momentum}a) and whose eigenvalue is the corresponding momentum timescale.

For some learned optimizers, this momentum eigenvalue exactly matched the best tuned momentum hyperparameter; in these cases the optimizer's performance matched that of momentum as well. We analyze one such optimizer in App.~\ref{appendix:momopt} as it is instructive for understanding the momentum mechanism.

\subsection{Gradient clipping}
\label{sec:clipping}

\begin{figure}[h]
\centering
\includegraphics[width=0.7\textwidth]{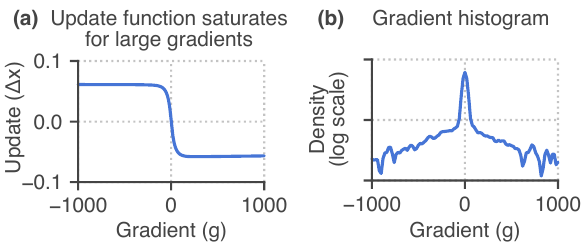}
\vspace{-1mm}
\caption{Gradient clipping in a learned optimizer trained on the Rosenbrock task (results for additional tasks are in App.~\ref{appendix:clipping}). \textbf{(a)}: The update function computed at the initial state saturates for large gradient magnitudes. The effect of this is similar to that of gradient clipping (cf. Fig.~\ref{fig:transfer}b). \textbf{(b)}: The empirical density of encountered gradients for this task. This shows that while most of the gradients occur in the linear regime, a small but non-negligible fraction are quite large and will saturate the update function.}
\label{fig:clipping}
\end{figure}

In standard gradient descent, the parameter update is a linear function of the gradient.
Gradient clipping \citep{pascanu2013difficulty} instead modifies the update to be a saturating function (Fig.~\ref{fig:transfer}b). This prevents large gradients from inducing large parameter changes, which is useful for optimization problems with non-smooth gradients~\citep{Zhang2020Why}.

We find that learned optimizers also use saturating update functions as the gradient magnitude increases, thus learning a soft form of gradient clipping.
We show this for the learned optimizer trained on the Rosenbrock problem in Figure \ref{fig:clipping}a.
Although Fig.~\ref{fig:clipping}a shows the saturation for a particular optimizer state (the initial state in this case), we find that these saturating thresholds are consistent throughout the optimizer state space. 

The strength of the clipping effect depends on the training task. We can see this by comparing the update function for a given optimizer to the distribution of gradients encountered for that task (Fig.~\ref{fig:clipping}b); the higher the probability of encountering a gradient that is in the saturating regime of the update function, the more clipping is used.

For some problems, such as linear regression, the learned optimizer largely stays within the linear region of the update function (App.~\ref{appendix:clipping}). For others, such as the Rosenbrock problem presented in Fig.~\ref{fig:clipping}, the optimizer utilizes more of the saturating part of the update function.

\subsection{Learning rate schedules}
\label{sec:schedules}

\begin{figure}[h]
\centering
\includegraphics[width=0.7\textwidth]{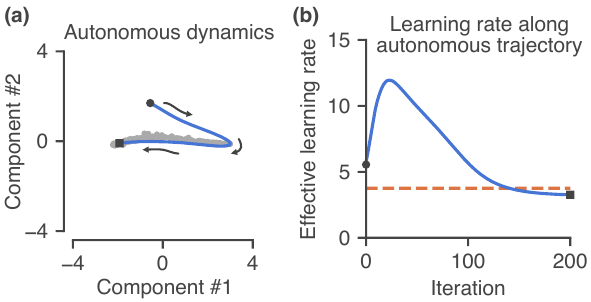}
\caption{Learning rate schedules mediated by autonomous dynamics, shown for the linear regression task (additional tasks are in App.~\ref{appendix:schedules}). \textbf{(a)}: Low-dimensional projection of the dynamics of the optimizer in response to no input (blue line) around approximate fixed points (gray circles). These autonomous dynamics allow the system to learn a learning rate schedule (see \S\ref{sec:schedules}). \textbf{(b)}: Effective learning rate computed at the autonomous state trajectories in (a). Dashed line indicates the best (tuned) learning rate for momentum on this task.}
\label{fig:schedules}
\end{figure}

Practitioners often tune learning rate schedules along with other optimization hyperparameters. Originally motivated to guarantee convergence in stochastic optimization \citep{robbins1951stochastic}, schedules are now used more broadly \citep{schaul2013no,smith2017don,ge2019step,choi2019empirical}. These schedules are typically a decaying function of the iteration\emdash{meaning the learning rate goes down as optimization progresses}although \citet{goyal2017accurate} use an additional increasing warm-up period, and even more exotic schedules have been proposed \citep{loshchilov2016sgdr,smith2017cyclical,li2019exponential}.

We discovered that learned optimizers can implement a schedule using \textit{autonomous}\emdash{that is, not input driven}dynamics.
If the initial optimizer state is away from fixed points of the state dynamics, then even in the absence of input, autonomous dynamics will encode a particular trajectory as the system relaxes to a fixed point.
This trajectory can
then be exploited 
by the learned optimizer to induce changes in optimization parameters, such as the effective learning rate.

Indeed, we find that for two of the tasks (linear regression and MNIST classification) learned optimizers learn an autonomous trajectory\footnote{Note that this autonomous trajectory evolves in a subspace orthogonal to the readout weights used to update the parameters.
This ensures that the autonomous dynamics themselves do not induce changes in the parameters, but only change the effective learning rate.} that modifies the learning rate, independent of being driven by any actual gradients.

This trajectory is shown for the linear regression task in Fig.~\ref{fig:schedules}a, starting from the initial state (black circle) and converging to a global fixed point (black square). Along this trajectory, we compute update functions and find that their slope changes; this is summarized in Fig.~\ref{fig:schedules}b as the effective learning rate changing over time. Results from the other tasks are presented in App.~\ref{appendix:schedules}.

\subsection{Learning rate adaptation}
\label{sec:adaptation}

\begin{figure*}[h]
\centering
\includegraphics[width=1.0\textwidth]{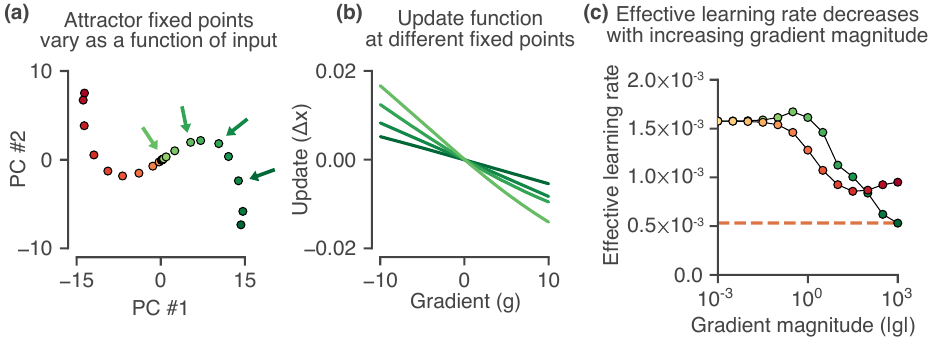}
\vspace{-2mm}
\caption{Learning rate adaptation in learned optimizers, shown for the Rosenbrock task (results are similar for other tasks, see App.~\ref{appendix:adaptation}). \textbf{(a)} Approximate fixed points (colored circles) of the optimizer state dynamics computed for different gradients reveals an S-curve structure. Large positive (negative) gradients push the optimizer state to the dark green (red) tails of the S-curve. \textbf{(b)} Update functions (\S\ref{sec:updatefunction}) computed at different points along the S-curve, corresponding to the arrows in (a). The effect of moving towards the tail of the S-curve is to make the update function more shallow (and thus have a smaller learning rate, cf.~Fig.~\ref{fig:transfer}d). The effect is similar along both arms; only one arm is shown for clarity. \textbf{(c)} Summary plot showing the effective learning rate along each arm of the S-curve, for negative (red) and positive (green) gradients.}
\label{fig:adaptation}
\end{figure*}

The next mechanism we discovered is a type of learning rate adaptation. The effect of this mechanism is to decrease the learning rate of the optimizer when large gradients are encountered. The effect is qualitatively similar to adaptive learning rate methods such as AdaGrad or RMSProp, but it is implemented in a new way in learned optimizers.

To understand how momentum is implemented by learned optimizers, we studied the linear dynamics of the optimizer near a fixed point (\S\ref{sec:momentum}). That fixed point was found numerically (\S\ref{sec:dynamicalsystems}) by searching for points $\vh^*$ that satisfy $\vh^* \approx F(\vh^*, g^*)$, where we hold the input (gradient) fixed at zero ($g^* = 0$). To understand learning rate adaptation, we need to study the dynamics around fixed points with non-zero input. We find these fixed points by setting $g^*$ to a fixed non-zero value.

We sweep the value of $g^*$ over the range of gradients encountered for a particular task. For each value, we find a single corresponding fixed point. These fixed points are arranged in an S-curve, shown in Figure \ref{fig:adaptation}a. The color of each point corresponds to the value of $g^*$ used to find that fixed point. One arm of this curve corresponds to negative gradients (red), while the other corresponds to positive gradients (green). 
The tails of the S-curve correspond to the largest magnitude gradients encountered by the optimizer, and the central spine of the S-curve contains the final convergence point\footnote{Fig.~\ref{fig:adaptation}a uses the same projection as in Fig.~\ref{fig:momentum}a, it is just zoomed out (note the different axes ranges).}.
 
These fixed points are all attractors, meaning that if we held the gradient fixed at a particular value, the hidden state dynamics would converge to that corresponding fixed point. In reality, the input (gradient) to the optimizer is constantly changing, but if a large (positive or negative) gradient is seen for a number of timesteps, the state will be attracted to the tails of the S-curve. As the gradient goes to zero, the system converges to the final convergence point in the central spine of Fig.~\ref{fig:adaptation}a.

What is the functional benefit of these additional dynamics? To understand this, we visualize the update function corresponding to different points along the S-curve (Fig.~\ref{fig:adaptation}b). The curves are shown for just one arm of the S-curve (green, corresponding to positive gradients) for visibility, but the effect is the symmetric across the other arm as well. We see that as we move along the tail of the S-curve (corresponding to large gradients) the slope of the update function becomes more shallow, thus the effect is to decrease the effective learning rate.

The learning rate along both arms of the S-curve are summarized in Fig.~\ref{fig:adaptation}c, for positive (green) and negative (red) gradients, plotted against the magnitude of the gradient on a log scale. This mechanism allows the system to increase its learning rate for smaller gradient magnitudes. For context, the best tuned learning rate for classical momentum
is shown as a dashed line.

\subsection{Tuning per layer and parameter type}
\label{sec:systemid}

\begin{figure*}[h]
\centering
\includegraphics[width=1.0\textwidth]{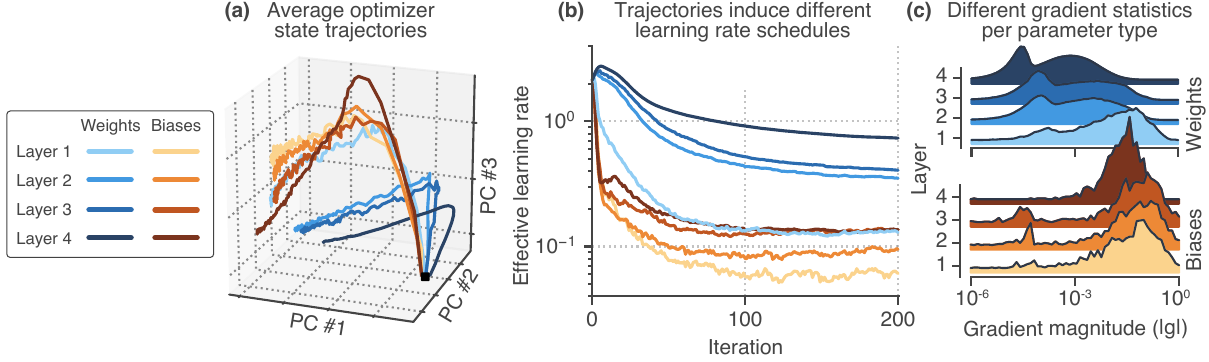}
\vspace{-3mm}
\caption{Parameter type-specific tuning found in a learned optimizer trained on an MNIST classification problem. \textbf{(a)} Trajectories of the optimizer, averaged across parameters within a particular layer and of a specific type (either weight or bias), exhibit different trajectories through optimizer state space. \textbf{(b)} Effective learning rate (computed as the slope of the update function) along each of the trajectories from panel (a). \textbf{(c)} Histograms of gradient magnitudes separated by layer and parameter type. The learned optimizer uses these differences in gradient statistics like these to induce the different trajectores from panels (a) and (b).}
\label{fig:systemid}
\end{figure*}
The final behavior we identified only exists in the learned optimizer trained on the MNIST CNN classification problem. It is a way for the learned optimizer to tune optimization properties (such as the effective learning rate) across different layers and across parameter types (either weight or bias) in the CNN being trained. We can see this most easily by taking all of the optimizer state trajectories for a particular layer or parameter type, and averaging them. These average trajectories are shown in Figure~\ref{fig:systemid}a projected onto the top three PCA components. The initial state is in the bottom right, and trajectories arc up and over before converging over to the left side of the figure. 

What is the functional benefit of separating out trajectories according to parameter type? To investigate this, we computed the update function at each point along the trajectories in Figure~\ref{fig:systemid}a. We found that the effective learning rate varied across them (Fig.~\ref{fig:systemid}b). In particular, the bias parameters all have a much lower learning rate than the weight parameters. Within a parameter type, later layers have larger learning rates than earlier layers.




A clue for how this happens can be found by looking at the gradient magnitudes across layers and parameter types at initialization (Fig.~\ref{fig:systemid}c). We see that the bias parameters (in orange) all have much larger gradient magnitudes on average compared to the weight parameters (in blue) in the later layers. This is a plausible hypothesis for the signal that the network uses to separate trajectories in Fig.~\ref{fig:systemid}a.

We want to emphasize that these are correlative, not causal, findings. That is, while the overall effect appears as if the network is separating out state trajectories based on layer or parameter type, it is possible that the network is really attempting to separate trajectories based on some additional factor that happens to be correlated with depth and or parameter type in this CNN.


\section{Discussion}
\label{discussion}

In this work, we trained learned optimizers on four different optimization tasks, and analyzed their behavior.
We discovered that learned optimizers learn a plethora of intuitive mechanisms: momentum, gradient clipping, schedules, forms of learning rate adaptation.
While the coarse behaviors are qualitatively similar across different tasks, the mechanisms are tuned for particular tasks.

While we have isolated specific mechanisms, we still lack a holistic picture of how these are stitched together.
One may be able to extract or distill a compressed optimizer from these mechanisms, perhaps using data-driven techniques~\citep{brunton2016discovering,champion2019data} or symbolic regression~\citep{cranmer2020discovering}.

The methods developed in this paper also pave the way for studies of when and how learned optimizers generalize.
By mapping different mechanisms to the underlying task used to train an optimizer, we can identify how quantitative properties of loss surfaces (e.g. curvature, convexity, etc.) give rise to particular mechanisms in learned optimizers.
Understanding these relationships would allow us to take learned optimizers trained in one setting, and know when and how to apply them to new problems.

Previously, not much was known about how learned optimizers worked.
The analysis presented here demonstrates that learned optimizers are capable of learning a number of interesting optimization phenomena.
The methods we have developed (visualizing update functions and linearizing state dynamics) should be part of a growing toolbox we can use to extract insight from the high-dimensional nonlinear dynamics of learned optimizers, and meta-learned algorithms more generally.

\subsubsection*{Acknowledgments}
The authors would like to thank C. Daniel Freeman and Ben Poole for helpful discussions and for comments on the manuscript.

\subsubsection*{Funding transparency}
The authors were employed by Google, Inc. while this research was being conducted.

\newpage

\bibliography{references}
\bibliographystyle{unsrtnat}

\appendix
\subsection*{Appendix for}
\section*{Reverse engineering learned optimizers reveals known and novel mechanisms}

\section{Comparing results across all mechanisms and tasks}
\label{appendix:results}

Below, we present and discuss results comparing mechanisms in learned optimizers across the different tasks studied (linear regression, Rosenbrock, Two Moons, and MNIST classification; see \S\ref{sec:methods}). This comparative anatomy across networks lets us see both how general particular mechanisms are, and hints at possible ways in which learned optimizers become tuned to a particular task distribution, at the expense of generalizing to other (untrained) tasks.

\subsection{Momentum}
\label{appendix:momentum}

\begin{figure*}[h!]
\centering
\includegraphics[width=0.85\textwidth]{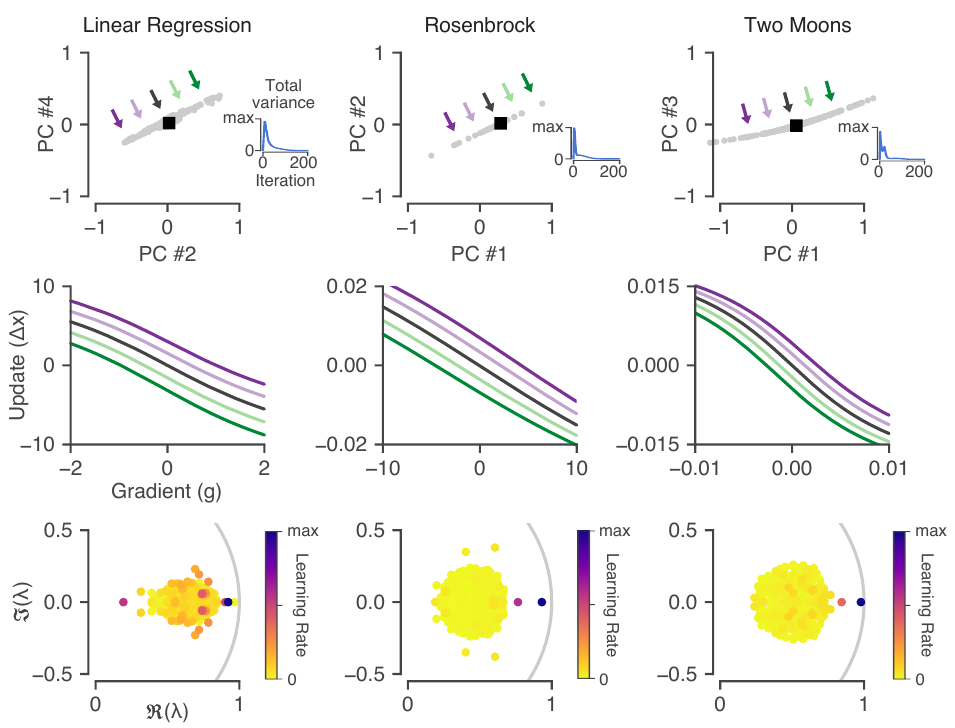}
\caption{Momentum in learned optimizers. Each column shows the same phenomena, but for optimizers trained on different tasks. \textbf{Top row}: Projection of the optimizer state around a convergence point (black square). {\it Inset:} the total variance of the optimizer states over test problems goes to zero as the trajectories converge. \textbf{Middle row}: visualization of the update functions (\S\ref{sec:updatefunction}) along the slow mode of the dynamics (colored lines correspond to arrows in (a)). Along this dimension, the effect on the system is to induce an offset in the update, just as in classical momentum (cf. Fig.~\ref{fig:transfer}c). \textbf{Bottom row}: Eigenvalues of the linearized optimizer dynamics at the convergence fixed point (black square in top row) plotted in the complex plane. The eigenvalue magnitudes are momentum timescales, and the color indicates the corresponding learning rate.}
\label{appfig:momentum}
\end{figure*}

We found highly consistent momentum mechanisms across the first three tasks studied (linear regression, Rosenbrock, and Two Moons classification). Results for these are shown in Figure~\ref{appfig:momentum}. Across these three tasks, we find that optimizer state trajectories converge to a single (global) fixed point (top row). This can be seen as the total variance across hidden states (inset in top row) goes to zero as training progresses. The dynamics around these final convergence points are organized around an approximate 1D line (indicated by gray dots in Fig.~\ref{appfig:momentum}). When moving along this line, the effect on the update function is to induce a vertical offset (Fig.~\ref{appfig:momentum}, middle row). Finally, the dynamics around these fixed points can be approximated using a linear approximation. This is done by computing the Jacobian of the system at the fixed point, and looking at its eigenvalues and eigenvectors. The bottom row of Figure~\ref{appfig:momentum} shows these eigenvalues in the complex plane. Across tasks, we find one or two modes that seem responsible for the dynamics. The eigenvalue of these modes can be interpreted as momentum timescales.

\begin{figure*}[h!]
\centering
\includegraphics[width=1.0\textwidth]{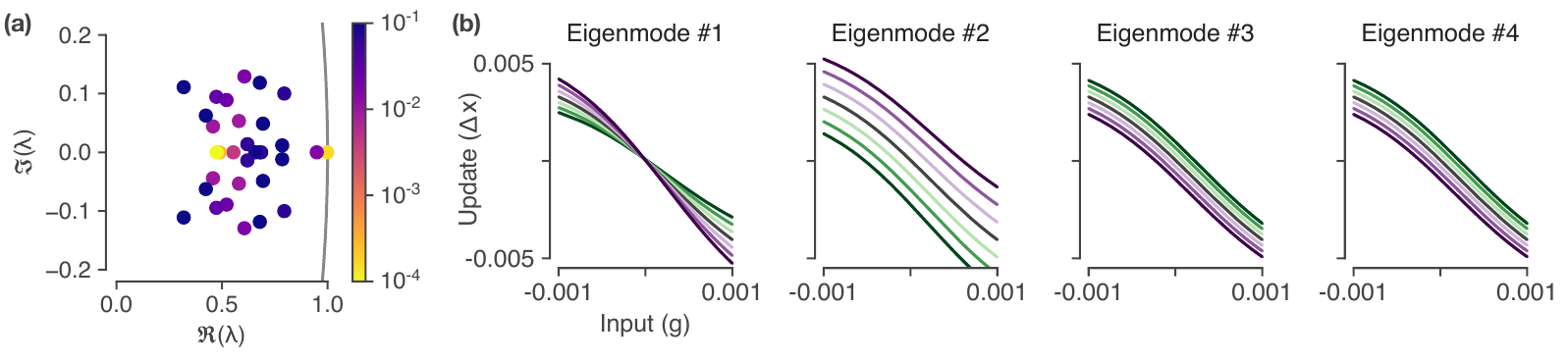}
\vspace{-5mm}
\caption{Momentum in a learned optimizer trained on MNIST. \textbf{(a)} Eigenvalues of the Jacobian of the optimizer dynamics, computed at an approximate convergence fixed point. There are multiple modes with large learning rates that seem to all contribute to the dynamics. \textbf{(b)} Update functions corresponding to moving along different eigenvectors of the Jacobian. For the first of these eigenmodes, the effect seems to be a change in the learning rate. For the rest, we see behavior that resembles momentum (a vertical shift in the update function).}
\label{appfig:mnistmom}
\end{figure*}

In a learned optimizer trained on MNIST, we found that trajectories did not converge to just one fixed point, but instead converged to any number of fixed points along a 1D manifold. The purpose of this manifold appears to be used for a type of learning rate adaptation, discussed more below in section \ref{appendix:adaptation}. What this means for the momentum mechanism, is twofold. First, the total variance of the trajectories no longer goes to zero, instead, it remains elevated as the trajectories for different parameters. Second, we find that there is not a single dominant eigenmode that seems responsible for momentum, the effect is spread out across multiple eigenmodes.

This is shown in Figure~\ref{appfig:mnistmom}. Fig.~\ref{appfig:mnistmom}a shows the eigenvalues of the Jacobian computed at one of the convergence fixed points along the manifold (others are similar). Instead of a single dominant mode, as observed for the other tasks, there are a number of modes with large learning rates. Fig.~\ref{appfig:mnistmom}b shows the effect of moving along the eigenvector associated with these top eigenvalues in state space. The first eigenmode corresponds to moving along the fixed point manifold\footnote{A oen dimensional fixed point manifold is an approximate line attractor; for a discrete time linear dynamical system this moving along the manifold corresponds to moving along an eigenvector whose corresponding eigenvalue is on the unit circle. Indeed, we see that the first eigenmode corresponds to the eigenvalue on the unit circle in Fig.~\ref{appfig:mnistmom}a.} and is discussed more below in the section on learning rate adaptation (\ref{appendix:adaptation}). However, eigenmodes 2-4 (and beyond) all have dynamics that look like momentum; our interpretation of this is that the network is using multiple momentum timescales as opposed to just one in this case.

\subsection{Gradient clipping}
\label{appendix:clipping}

\begin{figure*}[h!]
\centering
\includegraphics[width=1.0\textwidth]{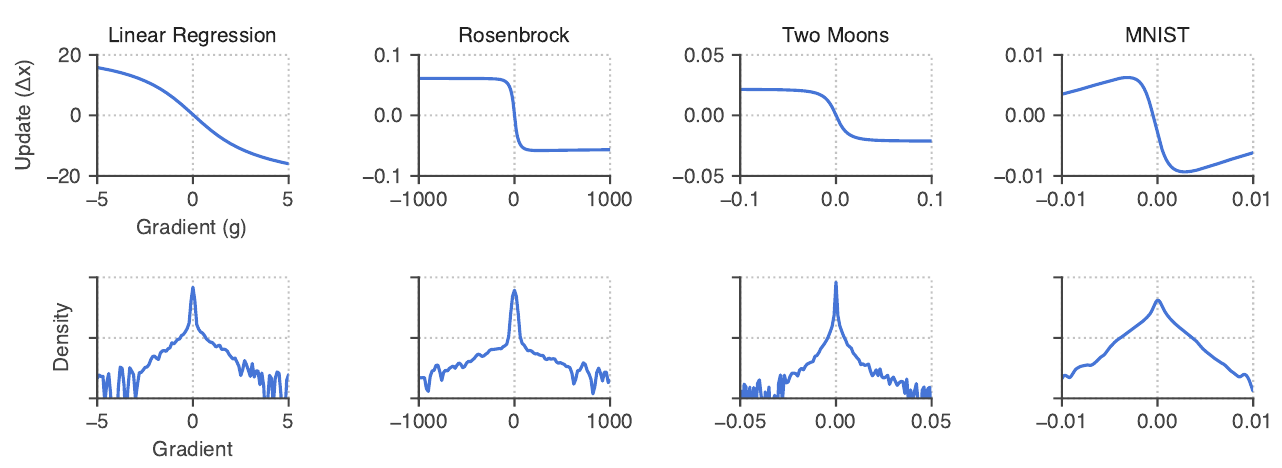}
\vspace{-5mm}
\caption{Gradient clipping in learned optimizers, trained on four different tasks. \textbf{Top row}: The update function computed at the initial state saturates for large gradient magnitudes. This effect is similar to that of gradient clipping (cf. Fig.~\ref{fig:transfer}b). \textbf{Bottom row}: the empirical density of encountered gradients for each task (note the different ranges along the x-axes). Depending on the problem, the learned optimizer can tune its update function so that most gradients are in the linear portion of the function, and thus not use gradient clipping (seen in linear regression, left column) or can potentially use more of the saturating region (seen on the Rosenbrock task, middle left).}
\label{appfig:clipping}
\end{figure*}

The gradient clipping effect (discussed in \S\ref{sec:clipping}) can be tuned by learned optimizers by making the update function more linear, or more saturating, over the range of expected gradient magnitudes. Indeed, we find this is the case for learned optimizers trained across the four tasks (Figure~\ref{appfig:clipping}). For some tasks, such as linear regression, the update functions look quite linear. For others, such as the Rosenbrock task, the network uses much more of the saturating part of the curve. Perhaps this is due to the large changes in curvature in the (non-convex) loss surface; clipping allows the network to have large learning rates in the valley of the Rosenbrock function, while not diverging if a step happens to take the network into the high curvature region outside the curved valley.

For the MNIST task (far right column of Fig.~\ref{appfig:clipping}), we find that the update function does not quite saturate, but instead slopes backwards, giving a non-monotonic update function. We found this consistently across multiple learned optimizers trained using different random seeds; however, the functional benefit of this non-monotonic behavior (if any) remains a mystery.

\subsection{Learning rate schedules}
\label{appendix:schedules}

\begin{figure*}[h!]
\centering
\includegraphics[width=1.0\textwidth]{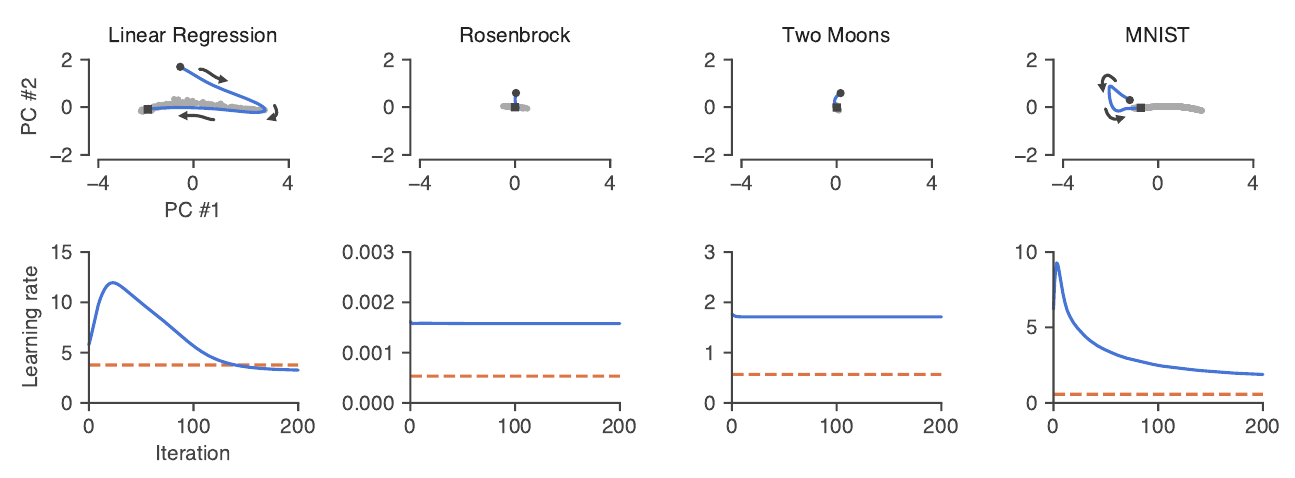}
\vspace{-5mm}
\caption{Learning rate schedules mediated by autonomous dynamics. \textbf{Top row}: Low-dimensional projection of the dynamics of the learned optimizer in response to zero gradients (no input). These autonomous dynamics allow the system to learn a learning rate schedule (see \S\ref{sec:schedules}). Gray circles are approximate fixed points of the dynamics. \textbf{Bottom row}: Effective learning rate (measured as the slope of the update function) as a function of iteration during the autonomous trajectories in the top row. We only observe a clear learning rate schedule in the linear regression task (far left) and the MNIST task (far right), both of which include a warm-up and decay. For context, dashed lines indicate the best (tuned) learning rate for momentum.}
\label{appfig:schedules}
\end{figure*}

Learned optimizers can learn an effective learning rate schedule through the use of autonomous (as opposed to input-driven) dynamics. The input to the learned optimizer is the gradient with respect to a particular parameter. When that derivative is zero, the network should not update the parameter. However, this does not mean that the hidden state needs to remain constant, instead, the hidden state is free to evolve along a subspace orthogonal to the readout used to update the parameter.

The functional benefit of this is that through the use of autonomous dynamics, the network can effectively change its behavior as a function of time (i.e.~the current optimization step). One type of behavioral change that we observe is a change in learning rate. This is prominent in learned optimizers trained on linear regression (far left column of Fig.~\ref{appfig:clipping}) and on MNIST (far right column of Fig.~\ref{appfig:clipping}).

\subsection{Learning rate adaptation}
\label{appendix:adaptation}

\begin{figure*}[h!]
\centering
\includegraphics[width=0.75\textwidth]{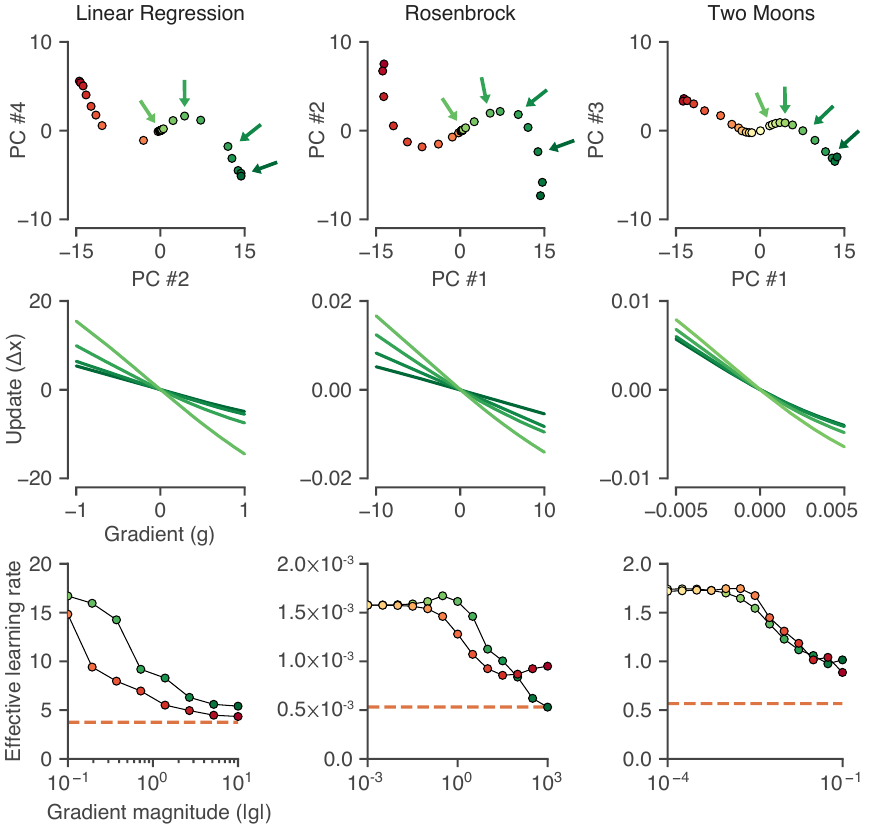}
\caption{Learning rate adaptation in learned optimizers. \textbf{Top row}: Approximate fixed points (colored circles) of the dynamics computed for different gradients reveal an S-curve structure. \textbf{Middle row}: Update functions (\S\ref{sec:updatefunction}) computed at different points along the S-curve (corresponding to arrows from the top row). The effect of moving towards the edge of the S-curve is to make the update function more shallow (thus have a smaller effective learning rate, cf.~Fig.~\ref{fig:transfer}d). The effect is similar along both arms; only one arm is shown for clarity. \textbf{Bottom row}: Summary plot showing the effective learning rate along each arm of the S-curve, for negative (red) and positive (green) gradients. The overall effect is to reduce learning rates when the gradient magnitude is large.}
\label{appfig:adaptation}
\end{figure*}

The final phenomenon we found across tasks is learning rate adaptation.
The overall effect of this phenomenon is to reduce (increase) the learning rate for parameters as their corresponding gradient magnitudes get large (small). While this effect is similar to that of RMSProp or Adam, we found the underlying mechanism in learned optimizers to be different.

For learned optimizers trained on the first three tasks, this mechanism is mediated by \textit{input-dependent} fixed points. These points are found by running the numerical optimization routine discussed in \S\ref{sec:dynamicalsystems}, but while holding the input fixed at various non-zero levels (rather than zero). The significance of this is that in the presence of large gradients, the network hidden state is away from the convergence point, and instead attracted towards different points (the input-dependent fixed points). The speed at which the hidden state approaches these fixed points depends on their dynamical properties, presumably this is something that can be tuned by the network as well. The functional benefit of this is that it allows the network to adapt its behavior \textit{after encountering large gradients for a number of consecutive iterations}.

Figure~\ref{appfig:adaptation} breaks down the mechanism found across the first three tasks. The top row of Fig.~\ref{appfig:adaptation} shows the set of input-dependent fixed points, which across these tasks forms an S-shaped curve. Large positive gradients (shown in green) pull the state along one arm of the S-curve, and large negative gradients (shown in red) pull the state along the opposite arm; the dynamics are symmetric across positive and negative gradients.

Along this curve, the network changes its behavior through a changing learning rate. We can see this as the slope of the update function changes (middle row of Fig.~\ref{appfig:adaptation}), these green lines correspond to the locations given by the arrows in the top row. Only changes along one arm of the S-curve are shown for clarity (for positive gradients), the effects are similar for negative gradients. The effect is shown for both arms in the bottom row of Fig.~\ref{appfig:adaptation}, summarized as the slope of the update function (the effective learning rate). Again, we see that the effect is to reduce the learning rate for large gradients. Interestingly, the most conservative learning rates for each task seem to match the best tuned learning rates used by momentum (shown as the dashed line). We hypothesize that momentum (whose learning rate does not adapt) must constrain its learning rate so that it does not diverge even in the off chance that a large gradient is encountered. Learned optimizers, on the other hand, are able to increase the learning rate as the gradient magnitudes decrease, which presumably partially explains their improved performance.

\begin{figure*}[h!]
\centering
\includegraphics[width=0.75\textwidth]{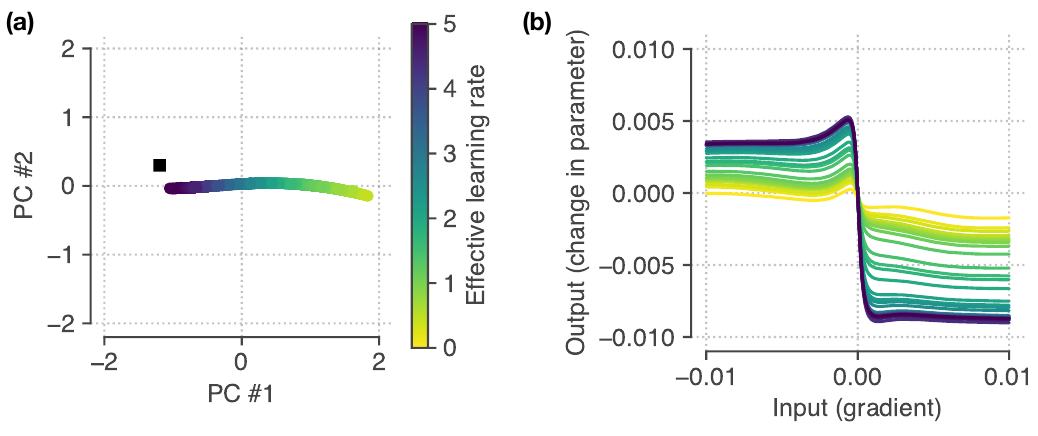}
\vspace{-3mm}
\caption{Variation in learning rate in a learned optimizer trained to optimize a network on MNIST classification. \textbf{(a)} Initial state (black square) and approximate fixed points (colored circles) found in a network trained on MNIST. Instead of a single fixed point, we see an approximately one dimensional manifold of fixed points. \textbf{(b)} Update functions corresponding to the points in the manifold in (a). Along this manifold, the overarching effect is a change in the effective learning rate.}
\label{appfig:mnistlr}
\end{figure*}

Finally, this was another mechanism where we see a different mechanism for the learned optimizer trained on MNIST classification. Here, we find that the network learns a manifold of fixed points (not just one) even when the input is held at zero. That is, while learned optimizers trained on the other tasks all converged to a single global fixed point, the MNIST networks converged to a 1D manifold of approximate fixed points. This manifold is shown in Figure~\ref{appfig:mnistlr}a. The functional benefit of this seems to be to induce a type of learning rate adaptation. Along this manifold, the update functions change dramatically (Fig.~\ref{appfig:mnistlr}b). One consequence of this is that the effective learning rate changes. This denoted by the variation in color in Fig.~\ref{appfig:mnistlr}, going from small learning rates (yellow) to large ones (purple). We suspect these fixed points are responsible for the dynamics that separates trajectories based on the underlying parameter type and layer (discussed in \S\ref{sec:systemid}).

\section{A learned optimizer that recovers momentum}
\label{appendix:momopt}

When training learned optimizers on the linear regression tasks, we noticed that we could train a learned optimizer that seemed to strongly mimic momentum, both in terms of behavior and performance. With additional training, the learned optimizer would eventually start to outperform momentum. However, it is still instructive to go through the analysis for the learned optimizer that mimics momentum. This example in particular clearly demonstrates the connections between eigenvalues, momentum, and dynamics.

\subsection{Recovering momentum using linear dynamics}

\begin{figure*}[h!]
\centering
\includegraphics[width=0.9\textwidth]{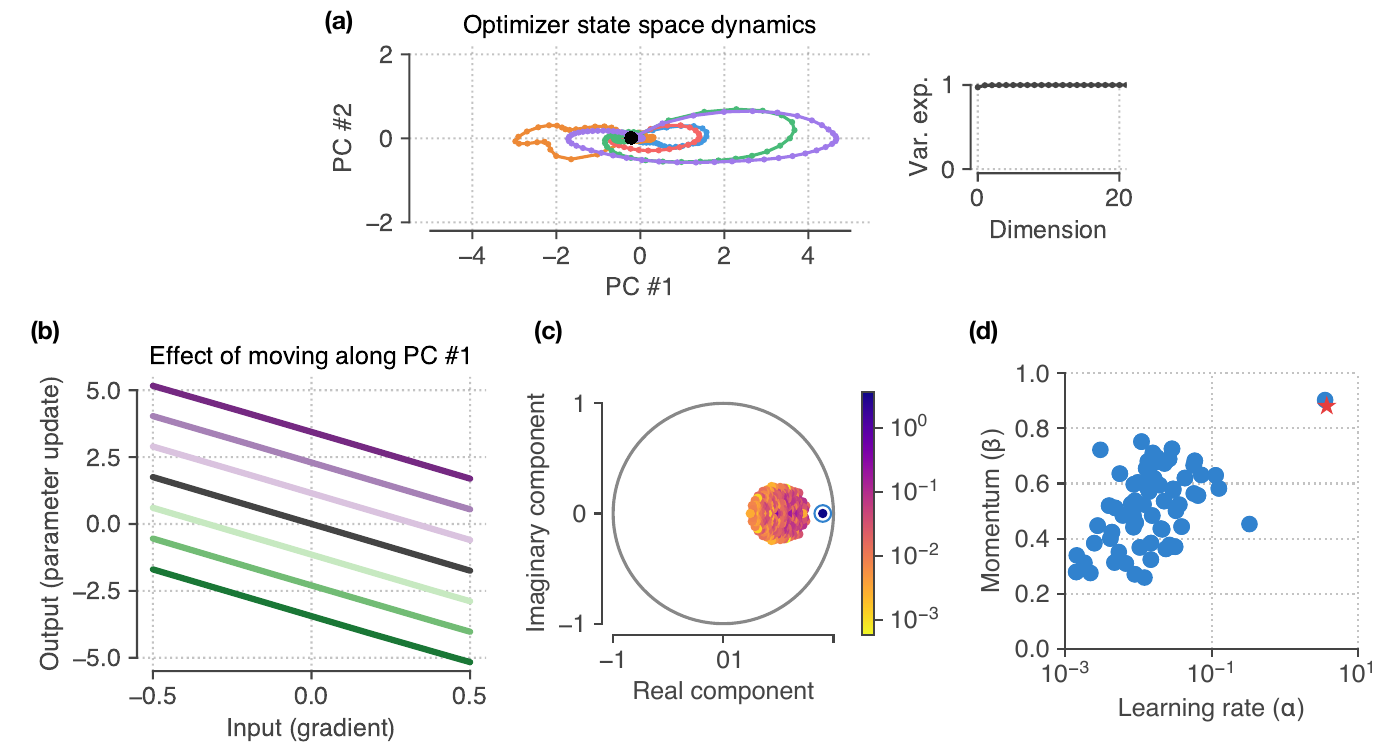}
\caption{\small A learned optimizer that recovers momentum on the linear regression task. \textbf{(a)} Optimizer state dynamics after training, projected onto the top principal components of the optimizer state space. Most of the action happens along a single dimension (PC 1). The dynamics converge to a single global fixed point (black circle). \textbf{(b)} Update functions of the optimizer along the first principal component. Moving along this component induces a vertical offset in the update function. \textbf{(c)} Eigenvalues of the Jacobian of the optimizer dynamics evaluated at the convergence fixed point. There is a single eigenmode that has separated from the bulk. The color of each point corresponds to the effective learning rate for that eigenmode. \textbf{(d)} Another way of visualizing eigenvalues is by translating them into optimization parameters (learning rates and momentum timescales), as described in section~\ref{appendix:aggmo}. When we do this for this particular optimizer, we see that the slow eigenvalue (momentum timescale closest to one) also has a large learning rate. These specific hyperparameters match the best tuned momentum hyperparametrs for this task distribution (red star).}
\label{fig:momopt}
\end{figure*}

The learned optimizer (parameterized by a GRU) that performs as well as momentum learns to mimic linear dynamics. That is, the dynamics of the nonlinear optimizer can be very well approximated using a linear approximation computed at the convergence point. The dynamics of this optimizer (projected onto the top principal components of the optimizer state space) are shown in Figure~\ref{fig:momopt}. A single principal component explains nearly all of the variance in hidden states (right inset of Fig.~\ref{fig:momopt}a). The optimizer state trajectories for an example problem are shown in Fig.~\ref{fig:momopt}a (recall that this problem is a five dimensional quadratic, so there are five trajectories). All trajectories converge to a single global fixed point, indicated by a black circle.

The update functions for this optimizer along the first principal component are shown in Fig.~\ref{fig:momopt}b (c.f.~Fig.~\ref{fig:transfer} and Fig.~\ref{fig:momentum}). The effect of moving along the first principal component looks exactly like what happens as you change the state variable in momentum. We can additionally analyze the dynamics of the optimizer hidden state by linearizing the nonlinear RNN dynamics around the fixed point. The eigenvalues of the Jacobian of the dynamics at the fixed point are shown in Fig.~\ref{fig:momopt}c. We find a single mode that pops out of the bulk of eigenvalues, indicated with a blue circle. The color in Fig.~\ref{fig:momopt}c corresponds to the effective learning rate of each eigenmode (as discussed below in section \ref{appendix:aggmo}).

Additionally, we can plot these eigenvalue magnitudes (which are the momentum time scales), against the extracted learning rate of each mode (Fig.~\ref{fig:momopt}d). The single mode that dominates the dynamics is in the upper right of the plot. Moreover, the extracted momentum timescale and learning rate for this mode match the best tuned hyperparameters (red star in Fig.~\ref{fig:momopt}d) from tuning the momentum algorithm directly, which can also be derived analytically.

Finally, if we extract and run just the dynamics along this particular mode, we see that it matches the behavior of the full, nonlinear optimizer almost exactly. This suggests that in this scenario, the learned optimizer has simply learned the single mechanism of momentum. Moreover, the learned optimizer has encoded the best hyperparameters for this particular task distribution in its dynamics. One benefit of our analysis is that we can now separate the overall mechanism (linear dynamics along eigenmodes) from the particular hyperparameters of that mechanism (the specific learning rate and momentum timescale).

\subsection{Meta-training dynamics}
\label{appendix:metatrainingdynamics}
\begin{figure*}[h!]
\centering
\includegraphics[width=0.9\textwidth]{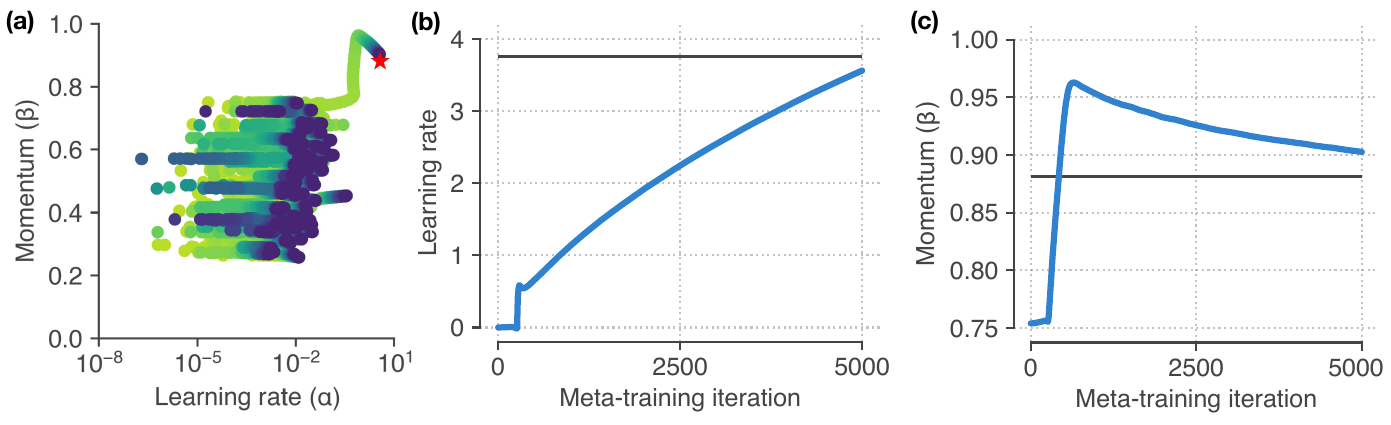}
\caption{\small Meta-training dynamics. \textbf{(a)} Meta-training dynamics of the learning rate and momentum parameters, extracted from the Jacobian of the convergence fixed point. Color indicates the meta-training iteration, from initialization (yellow-green) to when the optimizer is fully trained (purple). The red star indicates the optimal momentum parameters, derived analytically. \textbf{(b)} and \textbf{(c)} How the learning rate and momentum parameters for the top eigenmode evolve over the course of meta-training. Horizontal black line indicates the optimal value, derived analytically.}
\label{fig:mtd}
\end{figure*}

For this example, we also examined how the properties of the learned dynamics changed during meta-training. Specifically, we looked at how the effective learning rate and momentum of the optimizer (extracted from the Jacobian of the dynamics at the fixed point) varied over the course of meta-training (Figure~\ref{fig:mtd}). Fig.~\ref{fig:mtd}a shows the evolution of both the learning rate and momentum parameters for each eigenmode of the Jacobian. Panels (b) and (c) show the meta-training dynamics of just the top eigenmode (the only one that is used).

\section{Linearized optimizers and aggregated momentum}
\label{appendix:aggmo}

In this section, we elaborate on the connections between linearized optimizers and momentum with multiple timescales. We begin with our definition of an optimizer, equations (\ref{eq:dynamics}) and (\ref{eqn:readout}) in the main text:
\begin{eqnarray*}
    \vh^{k+1} &=& F(\vh^k, g^k) \\
    x^{k+1} &=& x^k + \vw^T \vh^{k+1},
\end{eqnarray*}
where $\vh$ is the optimizer state, $g$ is the gradient, $x$ is the parameter being optimized, and $k$ is the current iteration. Note that since this is a component-wise optimizer, it is applied to each parameter ($x_i$) of the target problem in parallel; therefore we drop the index ($i$) to reduce notation.

Near a fixed point of the dynamics, we approximate the recurrent dynamics with a linear approximation. The \textit{linearized} state update can be expressed as:
\begin{eqnarray}
    F(\vh^k, g^k) \approx \vh^* + \frac{\partial F}{\partial \vh} \left(\vh^k - \vh^*\right) + \frac{\partial F}{\partial g} g^k,
    \label{eq:lindyn}
\end{eqnarray}
where $\vh^*$ is a fixed point of the dynamics, $\frac{\partial F}{\partial \vh}$ is a square matrix known as the Jacobian, and $\frac{\partial F}{\partial g}$ is a vector that controls how the scalar gradient enters the system. Both of these latter two quantities are evaluated at the fixed point, $\vh^*$, and $g^*=0$.

For a linear dynamical system, as we have now, the dynamics decouple along eigenmodes of the system. We can see this by rewriting the state in terms of the left eigenvectors of the Jacobian matrix. Let $\vv = \mU^T \vh$ denote the transformed coordinates, in the left eigenvector basis $\mU$ (the columns of $\mU$ are left eigenvectors of the matrix $ \frac{\partial F}{\partial \vh}$). In terms of these coordinates, we have:
\begin{eqnarray}
    \vv^{k+1} = \vv^* + \mathbf{B} \left(\vv^k + \vv^* \right) + \va g^k,
    \label{eq:decoupled}
\end{eqnarray}
where $\mathbf{B}$ is a diagonal matrix containing the eigenvalues of the Jacobian, and $\va$ is a vector obtained by projecting the vector that multiplies the input $\left(\frac{\partial F}{\partial g}\right)$ from eqn.~(\ref{eq:lindyn}) onto the left eigenvector basis.

If we have an $N$-dimensional state vector $\vh$, then eqn.~(\ref{eq:decoupled}) defines $N$ independent (decoupled) scalar equations that govern the evolution of the dynamics along each eigenvector: $v_j^{k+1} = v_j^* + \beta_j \left(v_j^k + v_j^*\right) + \alpha_j g^k$,
where we use $\beta_j$ to denote the $j^{\text{th}}$ eigenvalue and $\alpha_j$ is the $j^{\text{th}}$ component of $\va$ in eqn.~(\ref{eq:decoupled}). Collecting constants yields the following simplified update:
\begin{eqnarray}
    v_j^{k+1} = \beta_j v_j^{k} + \alpha_j g + \text{const.},
    \label{eq:aggmo}
\end{eqnarray}
which is exactly equal to the momentum update ($v^{k+1} = \beta v^k + \alpha g^k$), up to a (fixed) additive constant. The main difference between momentum and the linearized momentum in eqn.~(\ref{eq:aggmo}) is that we now have $N$ different momentum timescales. Again these timescales are exactly the eigenvalues of the Jacobian matrix from above. Moreover, we also have a way of extracting the corresponding learning rate associated with eigenmode $j$, as $\alpha_j$. This particular optimizer (momentum with multiple timescales) has been proposed under the name \textit{aggregated momentum} by \citet{lucas2018aggregated}.

Taking a step back, we have drawn connections between a linearized approximation of a nonlinear optimizer, and a form of momentum with multiple timescales. What this now allows us to do is interpret the behavior of learned optimizers near fixed points through this new lens. In particular, we have a way of translating the parameters of a dynamical system (Jacobians, eigenvalues and eigenvectors) into more intuitive optimization parameters (learning rates and momentum timescales).

\section{Supplemental methods}
\label{appendix:suppmethods}

\subsection{Tasks for training learned optimizers}
\label{appendix:tasks}
An optimization problem is specified by both the loss function to minimize and the initial parameters.
When training a learned optimizer (or tuning baseline optimizers), we sample this loss function and initial condition from a distribution that defines a task.
Then, when evaluating an optimizer, we sample new optimization problems from this distribution to form a test set.

The idea is that the learned optimizer will discover useful strategies for optimizing the particular task it was trained on.
By studying the properties of optimizers trained across different tasks, we gain insight into how different types of tasks influence the learned algorithms that underlie the operation of the optimizer.
This sheds insight on the inductive bias of learned optimizers; i.e. we want to know what properties of tasks affect the resulting learned optimizer and whether those strategies are useful across problem domains.

We train and analyzed learned optimizers on three distinct tasks.
In order to train a learned optimizer, for each task, we must repeatedly initialize and run the corresponding optimization problem (resulting in thousands of optimization runs).
Therefore we focused on simple tasks that could be optimized within a couple hundred iterations, but still covered different types of loss surfaces: convex and non-convex functions, over low- and high-dimensional parameter spaces.
We also focused on deterministic functions (whose gradients are not stochastic), to reduce variability when training and analyzing optimizers.

\subsection{Training a learned optimizer}
\label{appendix:metatraining}

\begin{figure*}
\centering
\includegraphics[width=0.75\textwidth]{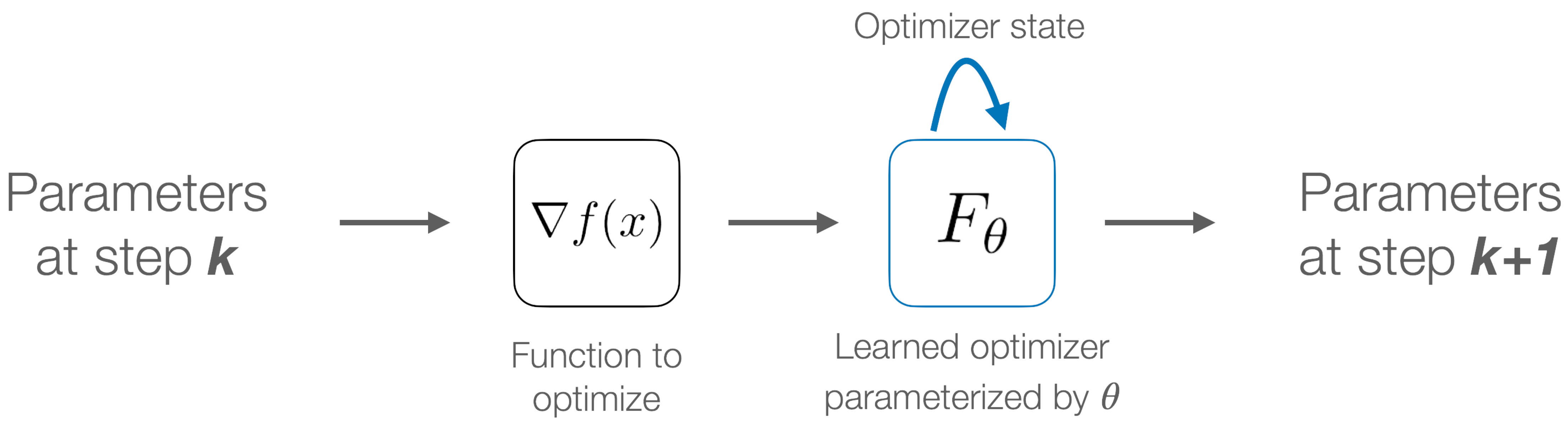}
\caption{Schematic of a learned optimizer.}
\label{fig:schematic}
\end{figure*}

We train learned optimizers that are parameterized by recurrent neural networks (RNNs). In all of the learned optimizers presented here, we use gated recurrent unit (GRU) \citep{cho2014learning} to parameterize the optimizer. This means that the function $F$ in eqn.~(\ref{eq:dynamics}) is the state update function of a GRU, and the optimizer state is the GRU state. In addition, for all of our experiments, we set the readout of the optimizer state, as defined in  eqn.~(\ref{eqn:readout}), to be linear. The parameters of the learned optimizer are now the GRU parameters, and the weights of the linear readout. We meta-learn these parameters through a meta-optimization procedure, described below.

In order to apply a learned optimizer, we sample an optimization problem from our task distribution, and iteratively feed in the current gradient and update the problem parameters, schematized in Figure~\ref{fig:schematic}. This iterative application of an optimizer builds an unrolled computational graph, where the number of nodes in the graph is proportional to the number of iterations of optimization (known as the length of the unroll). This is sometimes called the \textit{inner} optimization loop, to contrast it with the \textit{outer} loop that is used to update the optimizer parameters.

In order to train a learned optimizer, we first need to specify a target objective to minimize. In this work, we use the average loss over the unrolled (inner) loop as this meta-objective. In order to minimize the meta-objective, we compute the gradient of the meta-objective with respect to the optimizer weights. We do this by first running an unrolled computational graph, and then using backpropagation through the unrolled graph in order to compute the meta-gradient.

This unrolled procedure is computationally expensive. In order to get a single meta-gradient, we need to initialize, optimize, and then backpropagate back through an entire optimization problem. This is why we focus on small optimization problems, that are fast to train.

Another known difficulty with this kind of meta-optimization arises from the unrolled inner loop. In order to train optimizers on longer unrolled problems, previous studies have \textit{truncated} this inner computational graph, effectively only using pieces of it in order to compute meta-gradients.
While this saves computation, it is known that this induces bias in the resulting meta-gradients~\citep{wu2018understanding,metz2019understanding}.

To avoid this, we compute and backpropagate through fully unrolled inner computational graphs. This places a limit on the number of steps that we can then run the inner optimization for, in this work, we set this unroll length to 200 for all three tasks. Backpropagation through a single unrolled optimization run gives us a single (stochastic) meta-gradient, when meta-training, we average these over a batch size of 32.

Now that we have a procedure for computing meta-gradients, we can use these to iteratively update parameters of the learned optimizer (the outer loop, also known as meta-optimization). We do this using Adam as the meta-optimizer, with the default hyperparameters (except for the initial learning rate, which was tuned via random search). In addition, we use gradient clipping (with a clip value of five applied to each parameter independently and decay the learning rate exponentially (by a factor of 0.8 every 500 steps) during meta-training. We added a small $\ell_2$-regularization penalty to the parameters of the learned optimizer, with a penalty strength of $10^{-5}$. We trained each learned optimizer for a total of 5000 steps.

We meta-train our optimizers on a single TPUv2 core. Each model meta-trains in a few hours. Our code is built on top of the scientific python stack, including: NumPy~\citep{numpy}, SciPy~\citep{scipy}, Matplotlib~\citep{matplotlib}, and JAX~\citep{jax}.

For each task, we ended up with a single (best performing) learned optimizer architecture. These are the optimizers that we then analyzed, and form the basis of the results in the main text. The final meta-objective for each learned optimizer and best tuned baselines are compared below in Figure~\ref{fig:mobj}.

\begin{figure*}
    \centering
    \includegraphics[width=1.0\textwidth]{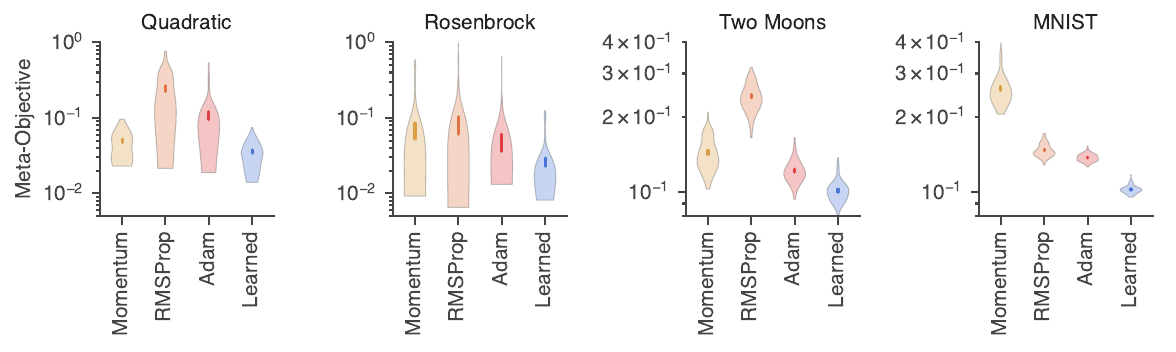}
    \vspace{-5mm}
    \caption{Performance summary. Each panel shows the distribution of the meta-objective over 64 random test problems for baseline and learned optimizers. Dark bars inside of each violin plot indicate the mean and standard error across the 64 random seeds. The learned optimizer has the lowest (best) meta-objective, on average, for each task.}
    \label{fig:mobj}
\end{figure*}

\subsection{Hyperparameter selection for baseline optimizers}
\label{appendix:baselines}

We tuned the hyperparameters of each baseline optimizer, separately for each task. For each combination of optimizer and task, we randomly sampled 2500 hyperparameter combinations from a grid, and selected the best one using the same meta-objective that was used for training the learned optimizer. We ensured that the best parameters did not occur along the edge of any grid.

\begin{figure*}
\centering
\includegraphics[width=0.8\textwidth]{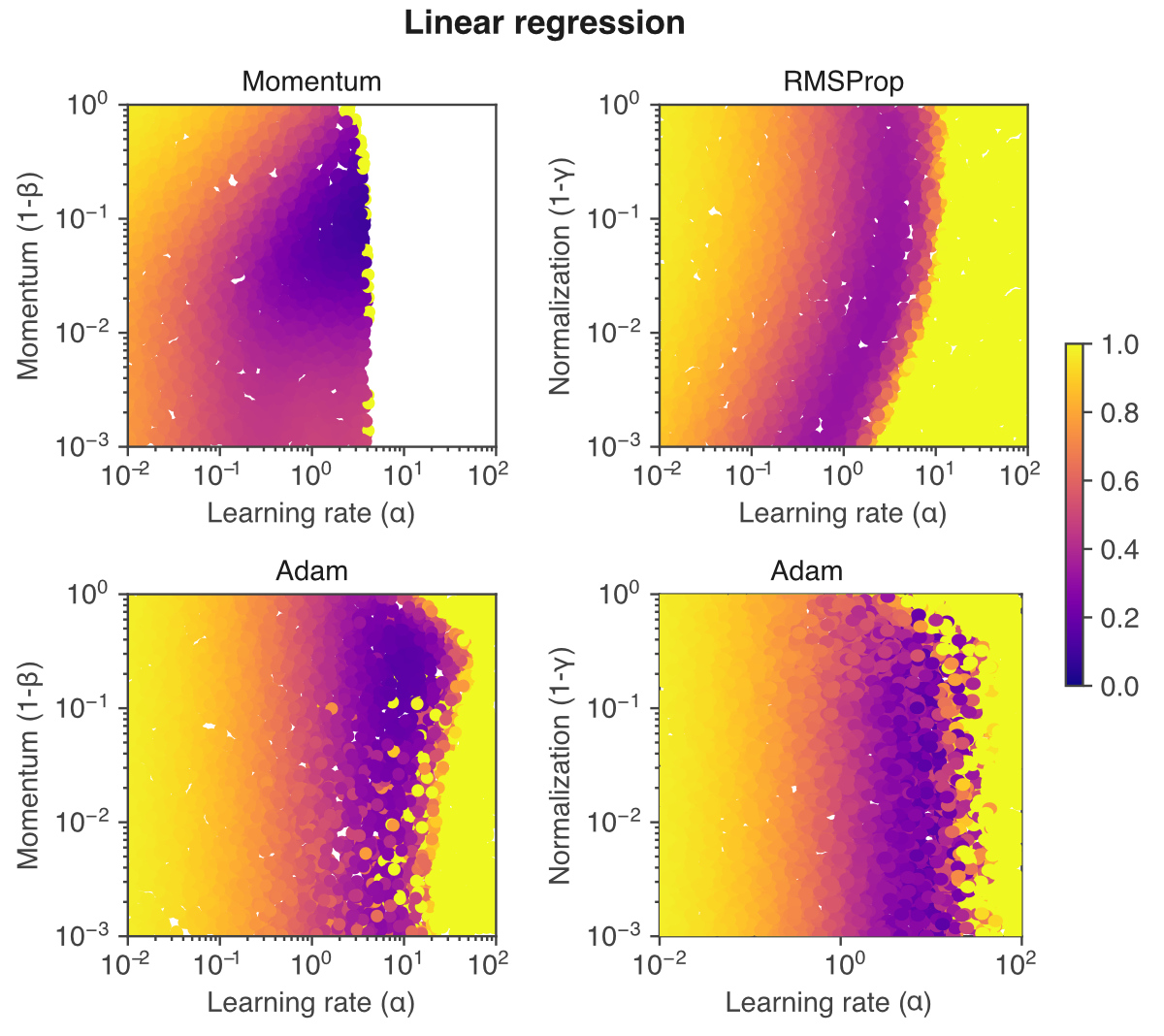}
\caption{Hyperparameter selection for linear regression.}
\label{fig:hypersquad}
\end{figure*}

\begin{figure*}
\centering
\includegraphics[width=0.8\textwidth]{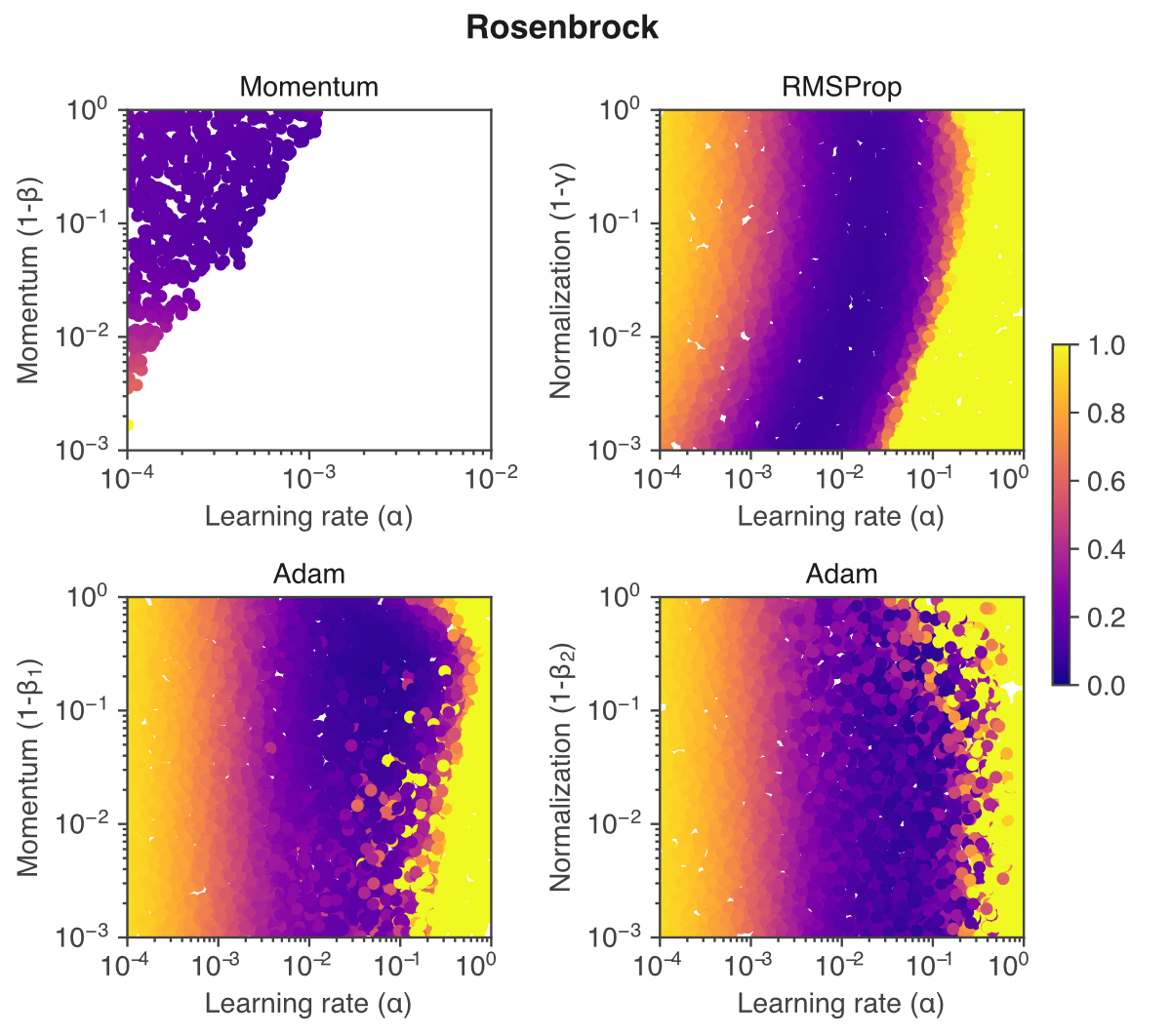}
\caption{Hyperparameter selection for Rosenbrock.}
\label{fig:hypersrosen}
\end{figure*}

\begin{figure*}
\centering
\includegraphics[width=0.8\textwidth]{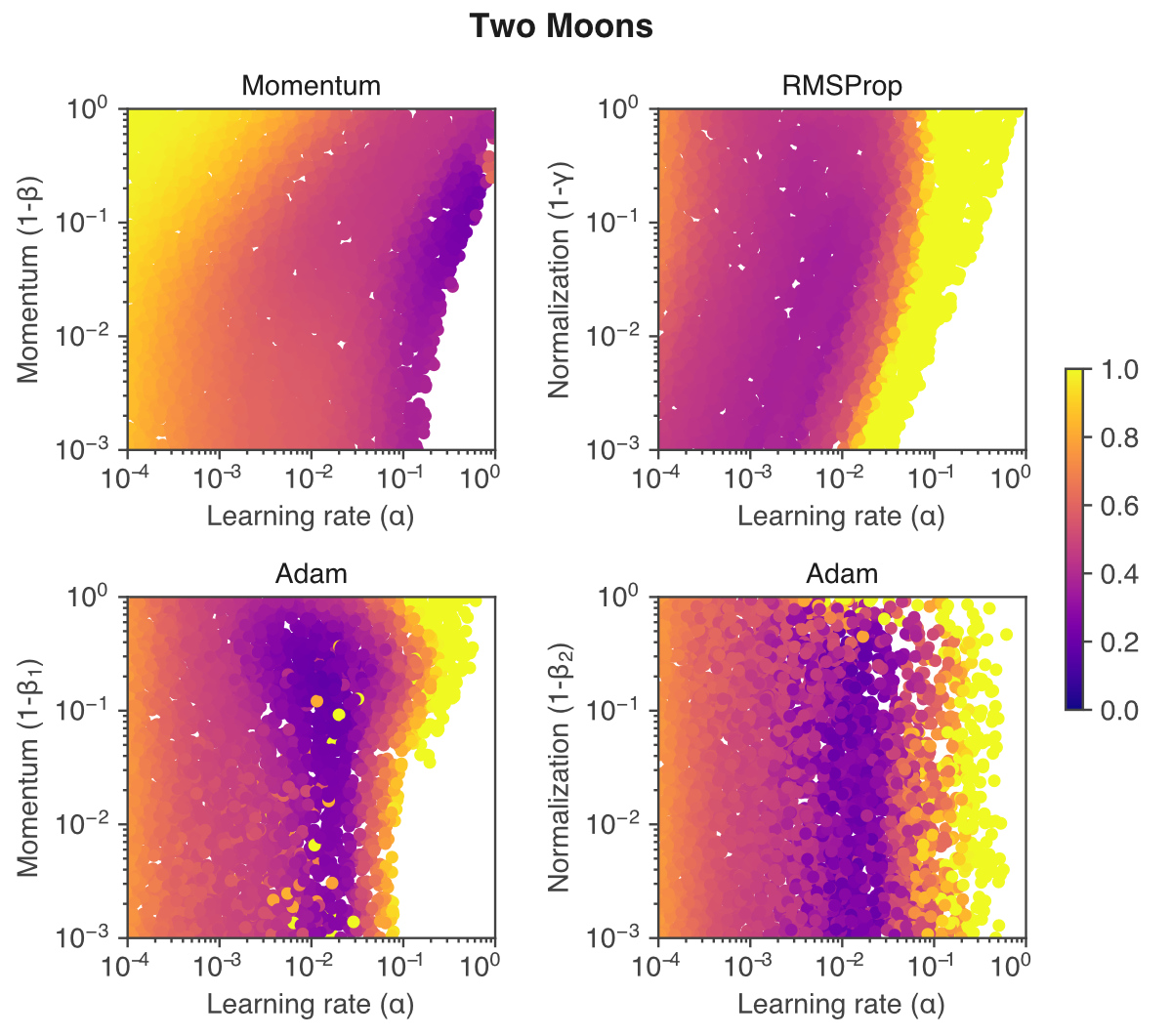}
\caption{Hyperparameter selection for training a neural network on two moons data.}
\label{fig:hypersmoons}
\end{figure*}

\begin{figure*}
\centering
\includegraphics[width=0.8\textwidth]{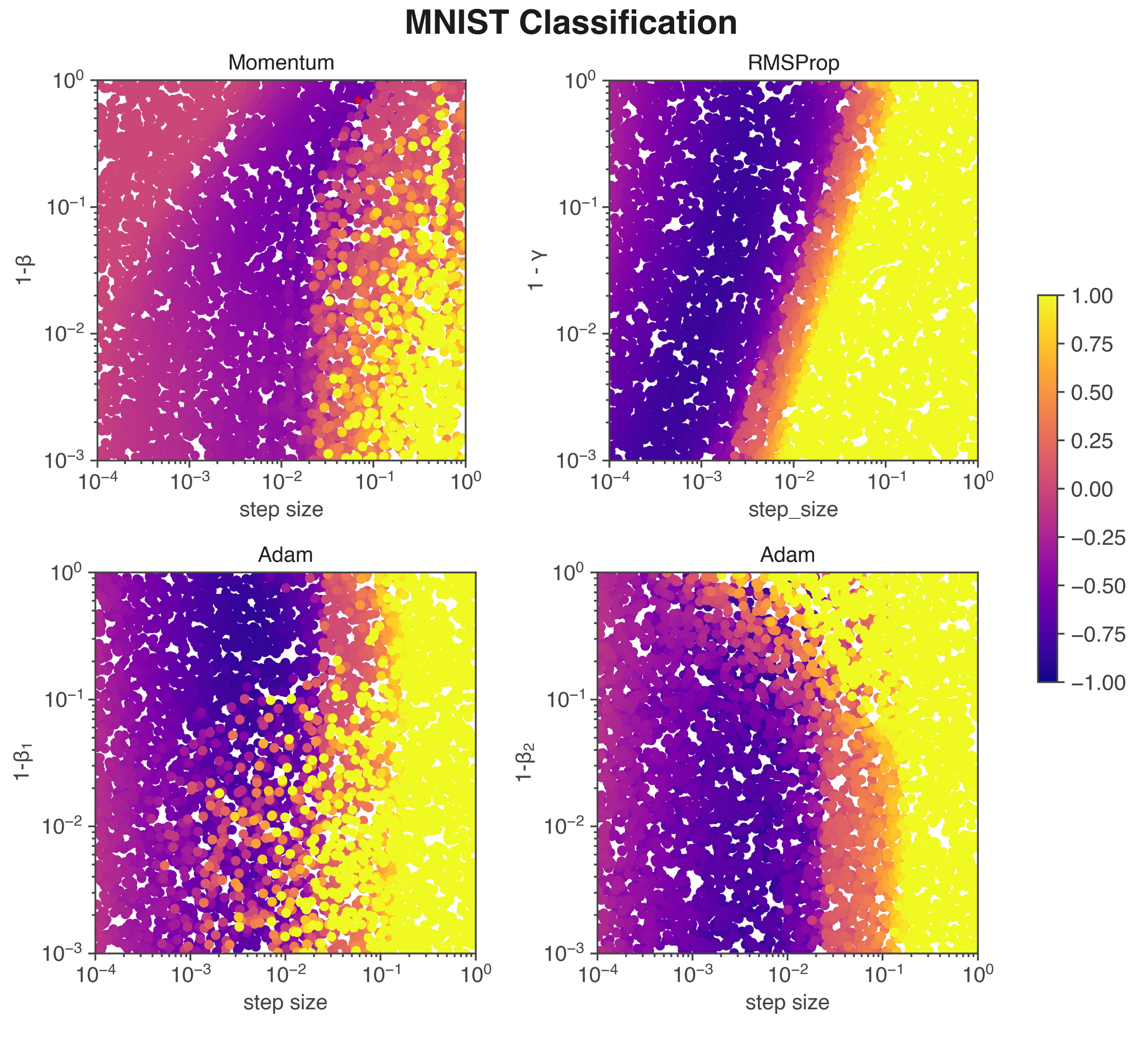}
\caption{Hyperparameter selection for training a neural network on MNIST classification.}
\label{fig:hypersmnist}
\end{figure*}

For momentum, we tuned the learning rate ($\alpha$) and momentum timescale ($\beta$). For RMSProp, we tuned the learning rate ($\alpha$) and learning rate adaptation parameter ($\gamma$). For Adam, we tuned the learning rate ($\alpha$), momentum ($\beta_1$), and learning rate adaptation ($\beta_2$) parameters. The result of these hyperparameter runs are shown in Figures \ref{fig:hypersquad} (linear regression), \ref{fig:hypersrosen} (Rosenbrock), \ref{fig:hypersmoons} (two moons classification), and \ref{fig:hypersmnist} (MNIST classification). In each of these figures, the color scale is the same\,---\,purple denotes the optimal hyperparameters.

\end{document}